\documentclass[12pt]{article}
\pdfoutput=1
\usepackage{amsmath}
\usepackage{graphicx}
\usepackage[backend=bibtex,style=authoryear,citestyle = authoryear, natbib = true, maxnames = 50,firstinits]{biblatex}
\usepackage{url} 

\usepackage{xcolor}

\newcommand{\blind}{0}

\renewcommand{\vec}[1]{\mathbf{#1}}
\DeclareMathOperator*{\argmax}{arg\,max}
\DeclareMathOperator*{\argmin}{arg\,min}

\usepackage{caption}
\usepackage{subcaption}
\usepackage[linesnumbered,lined,boxed,commentsnumbered]{algorithm2e}
\usepackage{mathtools}
\usepackage{amsfonts}
\usepackage{float}
\usepackage{graphicx}
\usepackage{amsthm}
\usepackage{array}
\usepackage{footnote}
\usepackage{tablefootnote}

\makeatletter
\renewenvironment{proof}[1][\proofname]{\par
  \pushQED{\qed}%
  \normalfont \topsep6\p@\@plus6\p@\relax
  \list{}{\leftmargin=4em
          \settowidth{\itemindent}{\itshape#1}%
          \labelwidth=\itemindent
          \listparindent=\parindent 
  }
  \item[\hskip\labelsep
        \itshape
    #1\@addpunct{.}]\ignorespaces
}{%
  \popQED\endlist\@endpefalse
}
\makeatother

\begin{document}

\def\spacingset#1{\renewcommand{\baselinestretch}%
{#1}\small\normalsize} \spacingset{1}


\if0\blind
{
  \title{\bf {VC-PCR}: A Prediction Method based on Supervised Variable Selection and Clustering}
  \author{Rebecca Marion\thanks{
    The authors gratefully acknowledge \textit{the Belgian Fund for Scientific Research (F.R.S.-FNRS, FRIA grant).}}\hspace{.2cm}\\
    Institute of Statistics, Biostatistics and Actuarial Sciences, \\
    LIDAM, UCLouvain, Belgium\\
    \\
    Johannes Lederer \\
    Department of Mathematics, Ruhr-University Bochum, Germany\\
    \\
    Bernadette Govaerts \\
    Institute of Statistics, Biostatistics and Actuarial Sciences, \\
   LIDAM, UCLouvain, Belgium\\
    \\
    Rainer von Sachs \\
    Institute of Statistics, Biostatistics and Actuarial Sciences, \\
    LIDAM, UCLouvain, Belgium\\}
    
  \maketitle
} \fi

\if1\blind
{
  \bigskip
  \bigskip
  \bigskip
  \begin{center}
    {\Large\bf {VC-PCR}: A Prediction Method based on Supervised Variable Selection and Clustering}
\end{center}
  \medskip
} \fi

\bigskip
\begin{abstract}

Sparse linear prediction methods suffer from decreased prediction accuracy when the predictor variables have cluster structure (e.g.\@ there are highly correlated groups of variables). To improve prediction accuracy, various methods have been proposed to identify variable clusters from the data and integrate cluster information into a sparse modeling process. But none of these methods achieve satisfactory performance for prediction, variable selection and variable clustering simultaneously. This paper presents Variable Cluster Principal Component Regression (VC-PCR), a prediction method that supervises variable selection and variable clustering in order to solve this problem. Experiments with real and simulated data demonstrate that, compared to competitor methods, VC-PCR achieves better prediction, variable selection and clustering performance when cluster structure is present.
\end{abstract}

\noindent%
\vfill

\newpage
\spacingset{1.5} 
\section{Introduction}
\label{sec:intro}

A common objective of transcriptomics is to predict health (cancer status, time to death, etc.) from genomic variables. For the sake of knowledge discovery and interpretability, it is also important to identify the variables that are most important for predicting the health variable (i.e. perform variable selection). However, for complex diseases, it is common for clusters of genes (i.e.\@ gene pathways) to contribute  to the disease jointly, rather than on an individual basis \citep{ma2007}. In these settings, the task of identifying clusters of predictive variables can be just as important as prediction and variable selection.

Transcriptomics data pose two major challenges to the tasks of prediction, variable selection and variable clustering. First, the data is high-dimensional, that is, there is a large number of variables (e.g.\@ genes) with respect to the number of observations. In this case, spurious correlations between relevant and irrelevant variables are common, and as a result, models tend to select too many variables. This overselection results in false positives \citep{daye2009}, i.e.\@ irrelevant variables that are identified as relevant. 
The second challenge stems from the underlying cluster structure among predictor variables: genes within a given pathway generally share similar functions and have highly correlated expression profiles. This high correlation between predictor variables is known to increase the variance of model coefficients, thereby decreasing prediction accuracy \citep{park2007clust}. 

To address these challenges, various methods have been proposed for estimating a sparse linear model that directly accounts for cluster structure among predictor variables. These methods mostly fall into one of two categories: embedded methods, which simultaneously cluster variables and estimate a sparse model, or two-step methods, which cluster variables and then estimate a sparse model based on these clusters. Embedded methods tend to be computationally expensive and are generally limited to regression settings, whereas two-step methods tend to identify clusters that are ``diluted'' with irrelevant variables \citep{yengo2014}. Methods from both categories tend to treat the task of variable clustering as a means to improve prediction accuracy rather than as an objective in itself. However, correctly identifying variable clusters (e.g. gene pathways in transcriptomics) is often instrumental to the interpretation of the model, as it 
allows researchers to better understand disease mechanisms, as well as develop targeted treatments.

\begin{figure}
    \centering
    \includegraphics[scale = 0.3]{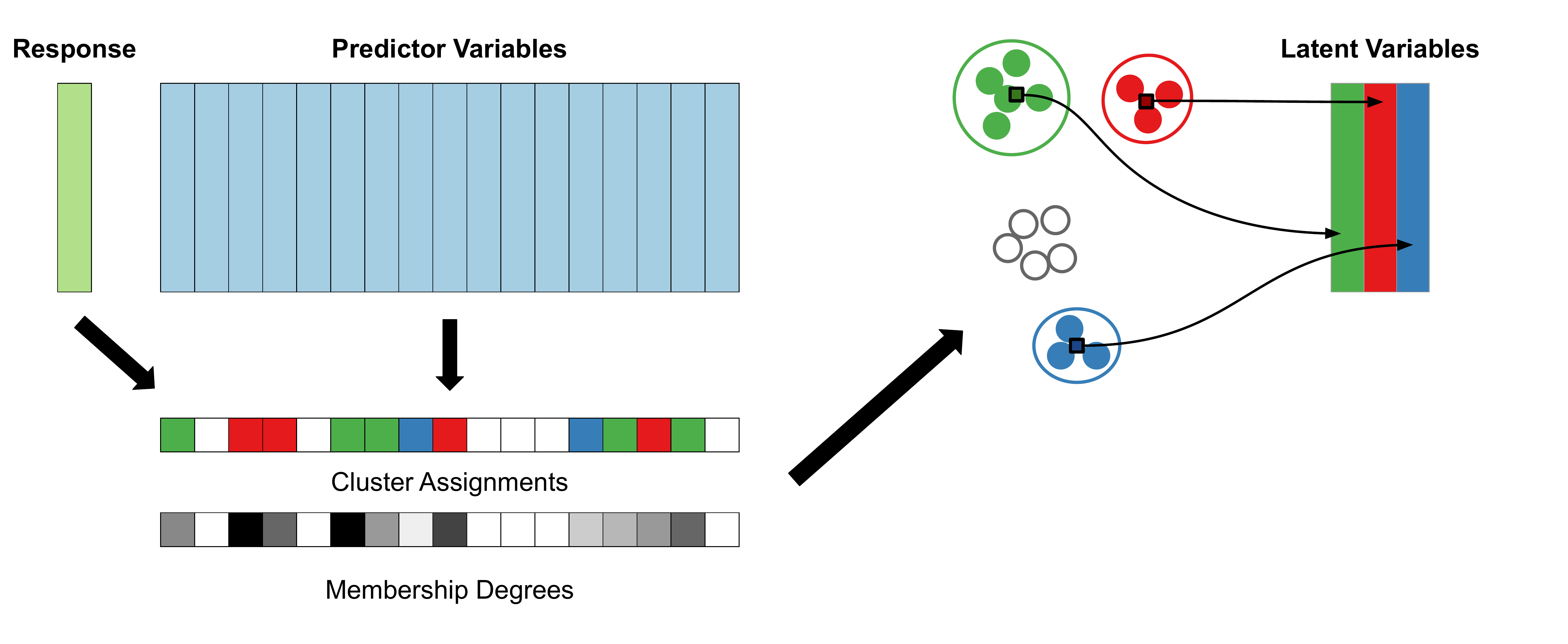}
    \\
    \includegraphics[scale = 0.3, trim = {0 0 20cm 0}, clip]{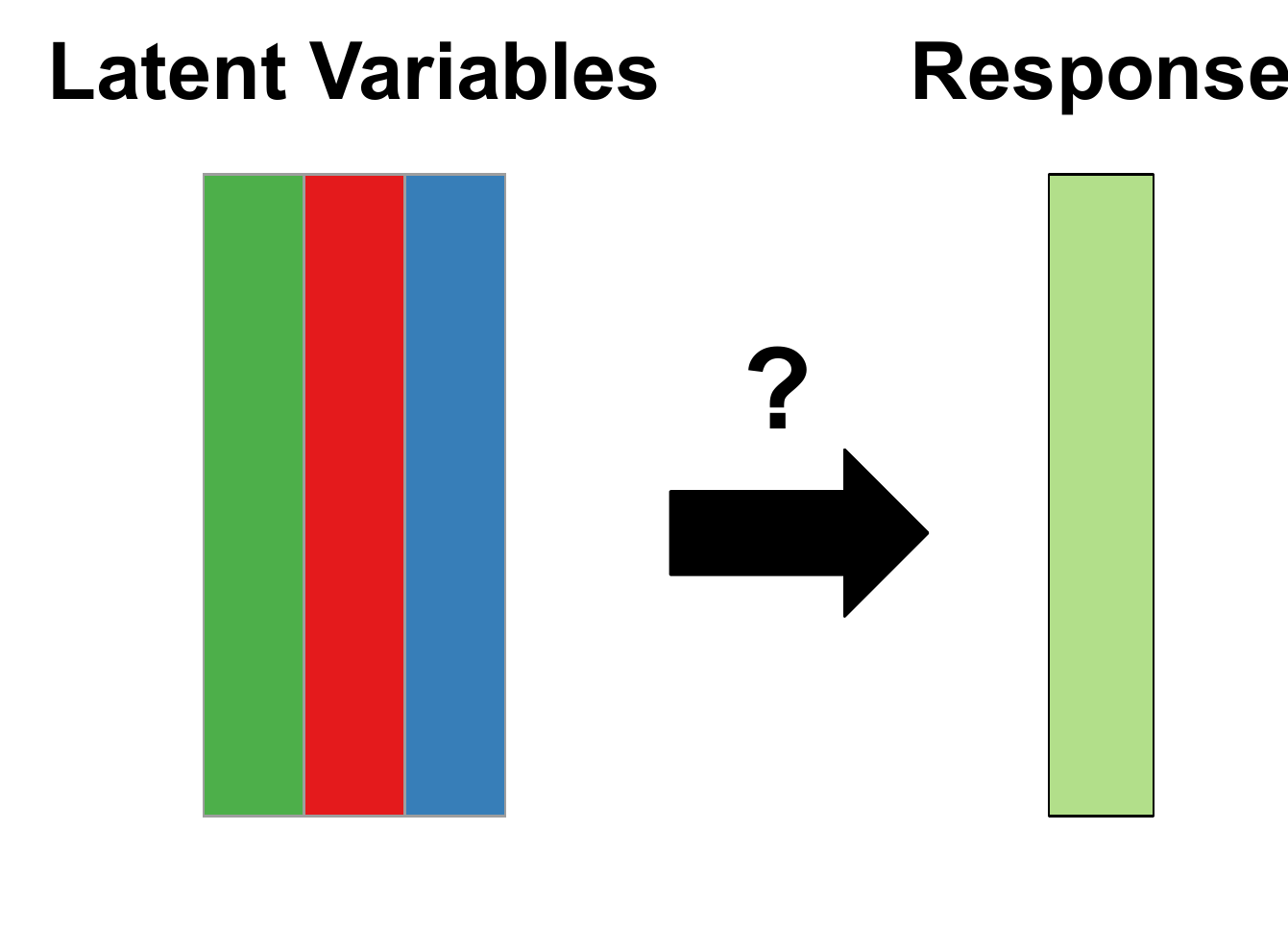}
    \caption{Schematic representation of VC-PCR. A response (green column) and predictors (light blue columns) are used to generate cluster assignments for the predictors. In the cluster assignments bar, predictors in the same cluster are shown in the same color, and predictors assigned to none of the clusters are shown in white. Membership degrees are also estimated for each predictor, with values ranging from no membership (white) to high membership (black). These membership degrees are used to calculate latent variables for each cluster. Then, a linear model is estimated based on these latent variables in order to predict the response.}
    \label{fig:VC-PCR_diagram}
\end{figure}

The goal of this paper is to address these limitations.
In this spirit, we propose a method called Variable Cluster Principal Component Regression (VC-PCR) that is motivated by the transcriptomics setup but is designed in a much more general fashion.
Unlike existing methods, it addresses three objectives simultaneously: (i) accurate prediction of the response in high-dimensional regression, (ii) selection of the most relevant predictor variables and (iii) identification of predictive variable clusters. The first step of VC-PCR consists of supervised variable selection and clustering, which is performed using a weighted version of a method that we call Sparse Orthogonal Semi-Nonnegative Matrix Factorization (SOS-NMF). As with the embedded approaches, supervising the variable clustering allows VC-PCR to identify homogeneous, predictive clusters. Moreover, irrelevant or ``inactive'' variables can be removed from the analysis during the clustering step, improving clustering performance and the recovery of the underlying model. In the second step of VC-PCR, a matrix of latent variables is calculated based on the cluster assignments from the first step, and a classical linear model is estimated to relate these latent variables to the response. Similar to existing two-step methods, the two-step approach employed by VC-PCR is more computationally efficient than embedded methods and can be easily applied to different prediction tasks (regression, classification, survival analysis, etc.). VC-PCR is summarized in a schematic representation in Figure~\ref{fig:VC-PCR_diagram}.

Numerical experiments with real and simulated data show that the proposed prediction method, VC-PCR, performs better than other competing methods in terms of model error, variable selection and clustering. VC-PCR with Ridge weights achieves the best variable selection and clustering performance across all simulation settings tested,  and  it  has  the  best  prediction  performance  for  sparse  models  in  settings  where variables have predictive cluster structure.

This paper is organized as follows. Section~\ref{sec:prelim} introduces the notation used in this paper. The proposed method, VC-PCR, as well as its optimization and properties, are presented in Section~\ref{sec:meth}. Related works are described in Section~\ref{sec:related}, and their limitations with respect to the current problem are discussed. Section~\ref{sec:exp} presents experiments comparing VC-PCR to competitor methods, where VC-PCR's superior performance is observed for both simulated and real datasets. Finally, Section~\ref{sec:concl} concludes the paper.

\section{Notation}
\label{sec:prelim}

Matrices are denoted with bold-faced capital letters (e.g.\@ $\vec{X}$, $\boldsymbol{\Theta}$), vectors with bold-faced lower-case letters (e.g.\@ $\vec{x}$, $\boldsymbol{\theta}$) and scalars with lower-case letters (e.g.\@ $x$, $\theta$). Row and column vectors of a matrix are indicated using subscripts. For example, if the columns of $\vec{X}$ are indexed by the letter $j$, then the $j$th column vector is denoted $\vec{x}_j$. If the rows are indexed by the letter $i$, then the $i$th row vector is denoted $\vec{x}_i$. A single element from a matrix is denoted using a lower-case letter with two subscripts, one for the row index and the other for the column index (e.g.\@ $x_{ij}$). Observations are indexed with the letter $i:1,...,n$, variables with the letter $j:1,...,p$ and clusters with the letter $k:1,...,K$. The Frobenius norm of an $n \times p$ matrix $\vec{X}$ is denoted $||\vec{X}||_F^2$, where $||\vec{X}||_F^2 = \sum_{i = 1}^n \sum_{j = 1}^p x_{ij}^2$. The $\ell_2$ norm of an $n \times 1$ vector $\vec{x}$ is $||\vec{x}||_2 = \sqrt{\sum_{i = 1}^n  x_{i}^2}$, and the $\ell_1$ norm of $\vec{x}$ is $||\vec{x}||_1 = \sum_{i = 1}^n |x_{i}|$. $\mathbb{I}\text{[} x \text{]}$ is an indicator function returning 1 if $x$ is true and 0 otherwise.

Let $\vec{X}$ ($n \times p$) be the matrix of predictor variables and $\vec{y}$ ($n \times 1$) be the response vector. We assume that the columns of $\vec{X}$ are mean-centered and have a sample variance of 1, i.e.\@ $\text{mean}(\vec{x}_j) = 0$ (sample mean), $\text{var}(\vec{x}_j) = 1$ (sample variance) and $\vec{x}_j^\top\vec{x}_j = n-1, \ \forall j$. For regression problems, the vector $\vec{y}$ is mean-centered and scaled to unit sample variance, and for two-class classification problems, the vector $\vec{y}$ is binary.

For regression problems, we assume that the response is a linear function of the predictor variables plus an error term:
\begin{equation}\label{eq:S1-linmod}
    y_i = \vec{x}_i^\top\boldsymbol{b} + \varepsilon_i,
\end{equation}
where $\boldsymbol{b}$ ($p \times 1$) contains the model coefficients relating the predictors to the response, $\vec{x}_i$ ($p \times 1$) contains the values of the predictors measured for observation $i$ and $\varepsilon_i \overset{{\text{\tiny iid}}}{\sim} N(0, \sigma_\varepsilon^2)$. For two-class classification problems, the elements in $\vec{y}$ are binary (e.g.\@ $y_i = 1$ if a patient has a given disease and $y_i = 0$ otherwise). We assume that $y_i \overset{{\text{\tiny iid}}}{\sim} Bernoulli(\pi_i)$, where
\begin{equation}\label{eq:S1-class}
     \pi_i = P(y_i = 1) = \frac{\exp( {\vec{x}}_i^\top\vec{b})}{1 + \exp( {\vec{x}}_i^\top\vec{b})}.
\end{equation}

\section{Variable Cluster -- PCR}
\label{sec:meth}

Our proposed method, Variable Cluster Principal Component Regression (VC-PCR), consists of two steps. In the first step, variables are clustered, generating a matrix $\widehat{\vec{V}}$ ($p \times K$) of cluster membership degrees. Each element $\widehat{v}_{jk}$ in $\widehat{\vec{V}}$ is nonnegative and represents the estimated degree to which variable $j$ belongs to cluster $k$. In the second step, $\widehat{\vec{V}}$ is used to calculate a set of latent variables, $\widehat{\vec{m}}_1, ..., \widehat{\vec{m}}_K$, one for each variable cluster, and a linear regression or classification model is estimated based on these latent variables. This second step is similar in spirit to Principal Component Regression (PCR), hence the name ``VC-PCR.''

The first step of VC-PCR (variable clustering) is performed using a method that we call Sparse Orthogonal Semi-Nonnegative Matrix Factorization (SOS-NMF). This method is summarized in Section~\ref{sec:S1-SOS-NMF}. The two steps of the VC-PCR algorithm are explained in Sections~\ref{sec:S1-step1} and \ref{sec:S1-step2}, and the full algorithm and a discussion of its hyperparameters are presented in Section~\ref{sec:S1-algo-hp}.

\subsection{Sparse Orthogonal Semi-NMF}\label{sec:S1-SOS-NMF}

Sparse Orthogonal Semi-Nonnegative Matrix Factorization (SOS-NMF) is the unsupervised method that we developed to perform dimensionality reduction, variable selection and variable clustering on both nonnegative matrices and matrices with positive and negative elements. The objective of SOS-NMF is to decompose $\vec{X}$ into the product of two matrices: $\vec{U}$ ($n \times K$), a matrix of $K$ unit-variance latent variables, and $\vec{V}$ ($p \times K$), a sparse, nonnegative matrix with orthogonal columns. Thanks to the orthogonality and nonnegativity of $\vec{V}$, each vector $\vec{v}_j$ (i.e. row of $\vec{V}$) can have at most one nonzero value. As a result, variable clusters can be inferred from $\vec{V}$: $C_k := \lbrace j \ | \ v_{jk} > 0 \rbrace$. Moreover, the nonzero elements of $\vec{V}$ indicate the degree to which a variable belongs to its assigned cluster, where larger values indicate larger membership degrees.

The solution to our proposed SOS-NMF problem is

\begin{equation}\label{eq:S1-SOS-NMF}
    \begin{aligned}
\argmin_{\vec{U}, \vec{V}} \ &  \frac{1}{2(n-1)} ||\vec{X} - \vec{U}\vec{V}^\top ||_F^2 + \lambda \sum_{j = 1}^p ||\vec{v}_j||_1\\  
 & \text{s.t. } v_{jk} \geq 0, \ \forall j,k\\
 & \phantom{\text{s.t. }} \text{var}(\vec{u}_k) = 1, \ \forall k\\
    & \phantom{\text{s.t. }} \vec{V} \text{ has orthogonal columns},\\
    \end{aligned}
\end{equation}
where $\lambda$ is a nonnegative hyperparameter controlling the sparsity in $\vec{V}$.

\subsubsection{Intuition}

The orthogonality and nonnegativity constraints make $\vec{V}$ a naturally sparse matrix. The $\ell_1$ penalty in Eq.~\eqref{eq:S1-SOS-NMF} encourages $\vec{V}$ to be even sparser, with some vectors $\vec{v}_j$ being set to zero (see, for example, Section~2.2 in \citep{lederer2021}). As a result, some variables (e.g.\@ noise variables) may be assigned to none of the clusters and removed from the analysis. This is desirable because noise variables make it difficult to correctly cluster variables (see \citet{vigneau2016,marion2020}).

A given variable $j$ is assigned to cluster $k$ (i.e.\@ $v_{jk} > 0$) if its correlation with latent variable $k$ is sufficiently large. As $\lambda$ increases, the correlation threshold for cluster assignment increases, resulting in the removal of more and more variables from the analysis. Another necessary condition for being assigned to cluster $k$ is having positive correlation with latent variable $k$. As a result, variables in the same cluster tend to be positively correlated.

Each latent variable vector $\vec{u}_k$ is proportional to the weighted sum of the variables $j$ assigned to cluster $k$, where the weights are the cluster membership degrees $v_{jk}$. This weighted sum makes it possible to give greater importance to the variables that are the most representative of the cluster.

\subsubsection{Optimization}
\label{sec:opt_SOS-NMF_intext}

The solution for SOS-NMF can be found by alternating between the optimization of $\vec{U}$ for fixed $\vec{V}$ and of $\vec{V}$ for fixed $\vec{U}$. The derivation of these solutions can be found in the Supplementary Material (Section~\ref{sec:opt_SOS-NMF}). For fixed $\vec{V}$, the solution for an unconstrained $\vec{U}$ is $\tilde{\vec{U}} = \vec{X}\vec{V}(\vec{V}^\top\vec{V})^{-1}$.  We can account for the unit-variance constraint on the latent variables $\vec{u}_k$ by calculating the final solution $\widehat{\vec{U}}$ as follows for all $k$:
\begin{equation}\label{eq:S1-scale_uk}
\widehat{\vec{u}}_k = \tilde{\vec{u}}_k / \text{sd}(\tilde{\vec{u}}_k).
\end{equation}
Let $C_k:= \lbrace j \ | \ v_{jk} > 0 \rbrace$. For fixed $\vec{U}$, if $j \in C_k$, then
\begin{equation}\label{eq:S1-vjk-SOS_intext}
     \widehat{v}_{jk} = \text{cor}(\vec{u}_k, \vec{x}_j) - \lambda. 
\end{equation}
The problem then becomes one of determining when $j \in C_k$. There are two necessary conditions for variable $j$ to belong to cluster $k$: (i) $\text{cor}(\vec{u}_k, \vec{x}_j) > \lambda$ and (ii) $ k = \argmax_{\ell} \text{cor}(\vec{u}_\ell, \vec{x}_j)$.

\subsection{VC-PCR: Step 1}\label{sec:S1-step1}

For the first step of VC-PCR, a matrix of cluster memberships $\vec{V}$ must be estimated. As one of our objectives is to predict a target $\vec{y}$, we propose making the clustering step dependent on the target. We do this by integrating information about the target into the SOS-NMF problem using a weight matrix $\vec{W}$. 

Let $\vec{W}$ ($p \times p$) be a diagonal matrix whose entries $w_{jj}$ encode (i) the prior relationship between variable $j$ and the target $\vec{y}$ (e.g.\@ positive vs negative association), as well as (ii) the importance of variable $j$ in the prediction task. For example, the weights could be the coefficients for a Ridge regression model (see Section~\ref{sec:S1-algo-hp} for other suggestions). When each variable $j$ is weighted by $w_{jj}$, the solution to SOS-NMF becomes

\begin{equation}\label{eq:S1-W-SOS-NMF}
    \begin{aligned}
\argmin_{\vec{U}, \vec{V}} \ &  \frac{1}{2(n-1)} ||\vec{X}\vec{W} - \vec{U}\vec{V}^\top ||_F^2 + \lambda \sum_{j = 1}^p ||\vec{v}_j||_1\\  
 & \text{s.t. } v_{jk} \geq 0, \ \forall j,k\\
 & \phantom{\text{s.t. }} \text{var}(\vec{u}_k) = 1, \ \forall k\\
    & \phantom{\text{s.t. }} \vec{V} \text{ has orthogonal columns},\\
    \end{aligned}
\end{equation}
where $\lambda$ is a nonnegative hyperparameter controlling the sparsity in $\vec{V}$. We call this problem Weighted SOS-NMF.

\subsubsection{Intuition}

The weight matrix $\vec{W}$ allows the clustering to be supervised by the target $\vec{y}$. For example, variables that have a similar importance for the prediction problem (i.e. similar weights in absolute value) are more likely to be clustered together. In addition, variables with smaller weights (in absolute value) are more likely to be removed by the $\ell_1$ penalty.

In contrast to the original SOS-NMF problem, variables may be assigned to cluster $k$ even if they are negatively correlated with latent variable $k$. Suppose there are two variables $j$ and $\ell$ such that $\text{cor}(\vec{u}_k,\vec{x}_j) = -\text{cor}(\vec{u}_k,\vec{x}_\ell) = \rho$ and $w_{jj} = -w_{\ell \ell} = \nu$. If  variable $j$ is assigned to cluster $k$, then variable $\ell$ will be as well. The higher the absolute correlation between variables, and the more similar their weights (in absolute value), the more likely it is that they will be clustered together.

\subsubsection{Optimization}

The solution for Weighted SOS-NMF can be found by using $\vec{X}\vec{W}$ as the input to SOS-NMF rather than $\vec{X}$. For fixed $\vec{V}$, the solution for an unconstrained $\vec{U}$ is $\tilde{\vec{U}} = \vec{X}\vec{W}\vec{V}(\vec{V}^\top\vec{V})^{-1}$. The solution for each unit-variance latent variable $\vec{u}_k$ can be calculated by scaling $\tilde{\vec{u}}_k$ (see Eq.~\eqref{eq:S1-scale_uk}). For fixed $\vec{U}$, a given variable $j$ is assigned to cluster $k$ if two conditions are met: (i) $w_{jj}\text{cor}(\vec{u}_k, \vec{x}_j) > \lambda$ and (ii) $k = \argmax_{\ell} w_{jj}\text{cor}(\vec{u}_\ell, \vec{x}_j)$. For all $j \in C_k$, the solution $\widehat{v}_{jk} =  w_{jj}\text{cor}(\vec{u}_k, \vec{x}_j) - \lambda$.

\subsection{VC-PCR: Step 2}\label{sec:S1-step2}

Once the matrix $\vec{V}$ of cluster memberships is estimated, the second step is to calculate a latent variable matrix $\widehat{\vec{M}} = f(\vec{X}, \widehat{\vec{V}})$ and estimate a linear model predicting $\vec{y}$ based on $\widehat{\vec{M}}$. Note that the latent variable matrix $\widehat{\vec{U}}$ from the first step is not used in the second step. This is because $\widehat{\vec{U}}$ is a function of $\vec{X}\vec{W}$, a modified version of $\vec{X}$. We are interested in relating the original matrix $\vec{X}$ to the target $\vec{y}$, therefore, the latent variable matrix $\widehat{\vec{M}}$ is calculated $\widehat{\vec{M}} = \vec{X}\widehat{\vec{V}}$.

The prediction problem requires solving

\begin{equation}\label{eq:S1-linprob}
    \argmin_{\vec{a}} \ L(\widehat{\vec{M}}, \vec{y} ; \vec{a}),
\end{equation}
where $\vec{a}$ ($K \times 1$) is a vector of model coefficients and $L(\widehat{\vec{M}}, \vec{y} ; \vec{a})$ is a loss function that depends on the modeling task. For linear regression tasks, the loss function $ L(\widehat{\vec{M}}, \vec{y} ; \vec{a}) = ||\vec{y} - \widehat{\vec{M}}\vec{a}||^2_2$ and $\vec{a}$ can be found using Ordinary Least Squares (OLS): $\widehat{\vec{a}} = (\widehat{\vec{M}}^\top \widehat{\vec{M}})^{-1} \widehat{\vec{M}}^\top \vec{y}$. Estimating a sparse regression model is not necessary because variable selection is already performed in the first step: variables are removed from the analysis thanks to the sparsity penalty in Eq.~\eqref{eq:S1-W-SOS-NMF}. For classification tasks with binary $y_i$, the likelihood-based loss function $L(\widehat{\vec{M}}, \vec{y} ; \vec{a}) = -\sum_{i = 1}^n y_i \log(\pi_i) + (1 - y_i)\log(1 - \pi_i)$, where 
\begin{equation}
   \pi_i = P(y_i = 1) = \frac{\exp(\widehat{\vec{m}}_i^\top\vec{a})}{1 + \exp(\widehat{\vec{m}}_i^\top\vec{a})}.
\end{equation}
The model coefficients $\vec{a}$ can be found using Logistic Regression. Again, sparse regression is not necessary here because variable selection has already been performed.

For both regression and classification tasks, a vector $\widehat{\boldsymbol{b}}$ ($p \times 1$) of coefficients for the original variables in $\vec{X}$ can be calculated as a function of $\widehat{\vec{V}}$ and $\widehat{\vec{a}}$. The predicted value $\widehat{y}_i$ is a function of $\widehat{\vec{m}}_i^\top\widehat{\vec{a}} = \vec{x}_i^\top\widehat{\vec{V}}\widehat{\vec{a}}$. Therefore, the coefficients $\widehat{\boldsymbol{b}}$ relating the original variables to the target can be calculated as $\widehat{\boldsymbol{b}} = \widehat{\vec{V}}\widehat{\vec{a}}$.

\subsection{Algorithm and Hyperparameters}\label{sec:S1-algo-hp}

This section summarizes the VC-PCR algorithm and offers guidance with respect to the choice of the hyperparameters $\lambda$ and $K$, as well as the weight matrix $\vec{W}$.

\subsubsection{Algorithm}

The entire VC-PCR algorithm is summarized below in Algorithm \ref{algo:VC-PCR}. At the beginning of the algorithm, $\vec{V}$ is initialized based on an initial partition $C_1, ..., C_K$ of $K$ variable clusters, where $v_{jk} = \mathbb{I}\left[j \in C_k \right]$. 

\begin{figure*}
\centering
\noindent
\resizebox{0.9\textwidth}{!}{%
\begin{minipage}{\textwidth}
\begin{algorithm}[H]
 \KwData{A predictor matrix ${\bf X}$ ($n \times p$) with mean-centered, unit-variance columns, a target vector $\vec{y}$ ($n \times 1$) (binary for classification, centered and scaled for regression), a diagonal weight matrix $\vec{W}$ ($p \times p$), an initial partition $C_1,...,C_K$ and a hyperparameter $\lambda > 0$}
 \tcp{Initialization}
 \For{each cluster $k$ in $1,...,K$}{
 \For{each variable $j$ in $1,...,p$}{
 $v_{jk} = \mathbb{I}\left[j \in C_k \right]$;
 	}
 }
 	
 	 \tcp{Step 1: Weighted SOS-NMF}
 \While{not converged}
 {
  \tcp{Update latent variable matrix}
  \If{there are columns of zeroes in $\vec{V}$}{
  Remove columns of zeroes from $\vec{V}$;
  }
  $\tilde{\vec{U}} = \vec{X}\vec{W}\vec{V}(\vec{V}^\top\vec{V})^{-1}$; \\
  
   \For{each cluster $k$ in $1,...,K$}
   {
     $\vec{u}_{ k} = \tilde{\vec{u}}_{ k}/\text{sd}(\tilde{\vec{u}}_{ k})$; }
 \tcp{Update cluster membership matrix} 
     \For{each variable $j$ in $1,...,p$}
   {
   $\widehat{k} = \argmax_{\ell} w_{jj}\text{cor}(\vec{u}_\ell, \vec{x}_j)$;\\
   $v_{j\widehat{k}} = (w_{jj}\text{cor}(\vec{u}_{\widehat{k}}, \vec{x}_j) - \lambda)_+$;
     
   }
   }
   
    $\widehat{\vec{V}} = \vec{V}$
 
  \tcp{Step 2: Linear Modeling}
  $\widehat{\vec{M}} = \vec{X}\widehat{\vec{V}}$;
  Calculate $\widehat{\vec{a}}$ by solving Eq.~\eqref{eq:S1-linprob};

\Return{Sparse cluster membership matrix $\widehat{\vec{V}}$, latent variable matrix $\widehat{\vec{M}}$ and model coefficients $\widehat{\vec{a}}$}
\caption{VC-PCR}
\label{algo:VC-PCR}
\end{algorithm}
\end{minipage}%
}%
\end{figure*}

\subsubsection{Choice of Weights}

As mentioned in Section~\ref{sec:S1-step1}, the entries $w_{jj}$ in $\vec{W}$ should indicate (i) the prior relationship between variable $j$ and the target $\vec{y}$ (e.g.\@ positive vs negative association) and (ii) the importance of variable $j$ in the prediction task. We propose two potential weighting schemes based on the coefficients of Lasso and Ridge models. Given $\vec{X}$, $\vec{y}$ and a hyperparameter $\delta$, a vector of coefficients $\vec{w}$ is estimated by solving

\begin{equation}\label{eq:S1-linprob_lasso}
    \argmin_{\vec{w}} \ L(\vec{X}, \vec{y} ; \vec{w}) + \text{Penalty}(\vec{w}; \delta),
\end{equation}
where $L(\vec{X}, \vec{y} ; \vec{w})$ is a loss function and $\text{Penalty}(\vec{w}; \delta)$ is a penalty function. For Lasso coefficients, this penalty is $\delta \sum_{j = 1}^p |w_{j}|$ and for Ridge coefficients, the penalty is $\delta \sum_{j = 1}^p w_{j}^2$.

When Lasso weights are used, prior variable selection is performed, as variables with Lasso coefficients equal to zero are essentially excluded from the variable clustering task. The remaining variables are weighted by a value that reflects their relative importance in the prediction task, as well as the direction of their correlation with the target.  Ridge weights encode similar information but do not pre-select variables: in most cases, all variables are given a nonzero weight. However, variables with high absolute correlation are more likely to have similar weights in absolute value. This encourages SOS-NMF to cluster these variables together.

\subsubsection{Choice of Hyperparameters}

The VC-PCR algorithm depends on three hyperparameters: $K$, the number of clusters in the initial partitions, $\delta$, the hyperparameter used to estimate the weights in $\vec{W}$, and  $\lambda$, which determines the degree of sparsity in $\vec{V}$. In practice, these hyperparameters are tuned using $k$-fold cross-validation, using a criterion such as the Mean Squared Error of Prediction (MSEP) for regression problems or the Matthews Correlation Coefficient (MCC) \citep{matthews1975} for classification problems.

For the sparsity hyperparameter $\lambda$, the smallest value leading to $\vec{V} = \vec{0}$ can be determined based on the first iteration of the while loop in Algorithm \ref{algo:VC-PCR}. Let $\vec{u}_k^{(1)}$ be the $k^{th}$ latent variable at the first iteration. All elements $v_{jk}$ are equal to zero if $\lambda$ is greater than or equal to $w_{jj}\text{cor}(\vec{u}_k^{(1)}, \vec{x}_j)$, for all $j,k$. Therefore, the largest value of $\lambda$ to consider is
\begin{equation}
    \lambda_{\max} = \max_{j,k} \ w_{jj}\text{cor}(\vec{u}_k^{(1)}, \vec{x}_j).
\end{equation}

\section{Related Work}\label{sec:related}

In this section, methods related to the proposed SOS-NMF and VC-PCR methods are presented. In Section~\ref{sec:S1-NMF}, several nonnegative matrix factorization (NMF) methods are summarized and compared to SOS-NMF. In Sections~\ref{sec:S1-coeff_group}-\ref{sec:S1-two_step}, three different classes of prediction methods are described and compared to VC-PCR: coefficient-grouping methods (Section~\ref{sec:S1-coeff_group}), ensemble methods (Section~\ref{sec:S1-ensemble}) and two-step methods (Section~\ref{sec:S1-two_step}).

\subsection{Nonnegative Matrix Factorization Methods}\label{sec:S1-NMF}

SOS-NMF is most closely related to the problem of Nonnegative Matrix Factorization (NMF). While certain methods exist for addressing aspects of the SOS-NMF problem, no method satisfactorily addresses all aspects simultaneously (i.e.\@ sparsity, orthogonality and semi-nonnegativity). This section explores the relationship between SOS-NMF and several closely related NMF methods from the literature. Comparisons to other NMF methods (Semi-NMF \citep{ding2008}, Orthogonal NMF \citep{ding2006} and Sparse NMF \citep{kim2007SNMF}) can be found in the Supplementary Material (Section~\ref{sec:otherNMF}). The objective functions and constraints associated with the NMF methods studied in this paper are summarized in Table~\ref{tab:S1-NMF_methods}.

Sparse Orthogonal NMF (SONMF) \citep{dai2018} constrains $\vec{V}$ to be sparse, nonnegative and orthogonal.  In contrast to SOS-NMF, however, the input $\vec{Z}$ to SONMF must be nonnegative, making SONMF less flexible. Moreover, the scale of the columns of $\vec{V}$ is constrained, rather than the scale of the columns of $\vec{U}$.  These scaling differences have implications for the estimation of $\vec{V}$. For SOS-NMF, the solution for $\vec{v}_j$ can be solved independently for each variable $j$ because the scale of each vector $\vec{v}_k$ is not fixed. For SONMF, the solution for $v_{jk}$ depends on all other entries in $\vec{V}$ due to the orthonormality constraint. As a result, the solution for each vector $\vec{v}_j$ cannot be found independently, and the scale of $v_{jk}$ for variables $j \in C_k$ depends on the size of cluster $k$. The more variables that are assigned to cluster $k$, the more likely it is that these variables will have low memberships $v_{jk}$, to satisfy the constraint $\vec{v}_k^\top\vec{v}_k = 1$. This unfairly penalizes large clusters, as the sparsity penalty makes small values $v_{jk}$ shrink to zero.

SOS-NMF is also related to Nonnegative Sparse Principal Component Analysis \citep{zass2007SNPCA}. In contrast to SOS-NMF, the columns of $\vec{V}$ in NSPCA are only quasi-orthogonal, unless $\gamma = \infty$. This means that variables may be assigned to multiple clusters (i.e.\@ $\vec{v}_j$ may have multiple nonzero entries). For the special case where $\gamma = \infty$, NSPCA only differs from SOS-NMF with respect to the scaling constraints on $\vec{U}$ and $\vec{V}$: SOS-NMF constrains the scale of $\vec{U}$ whereas NSPCA constrains the scale of $\vec{V}$. This is demonstrated in further detail in the Supplementary Material (Section~\ref{sec:S1-orthoNSPCA_proof}). To the best of our knowledge, no optimization strategies have been proposed for solving the NSPCA problem when $\gamma = \infty$, referred to here as Orthonormal NSPCA. Moreoever, Orthonormal NSPCA suffers from the same scaling problems as Sparse Orthogonal NMF (SONMF): the scaling constraints make the estimation of $\vec{v}_j$ dependent on all of the other variables $\ell \neq j$.

\begin{table}[ht]
\resizebox{\textwidth}{!}{%
    \begin{tabular}{lll}
    \textbf{Method} & \textbf{Objective Function} & \textbf{Constraints} \\ \hline
        Semi-NMF\tablefootnote{\cite{ding2008}} & $\frac{1}{2(n-1)} ||\vec{X} - \vec{U}\vec{V}^\top ||_F^2$ & $v_{jk} \geq 0, \ \forall j,k$ \\ \hline
        Orthogonal NMF\tablefootnote{\cite{ding2006}} & $\frac{1}{2(n-1)} ||\vec{Z} - \vec{U}\vec{V}^\top ||_F^2$ & $v_{jk} \geq 0, \ \forall j,k$ \\
& & $u_{ik} \geq 0, \ \forall i,k$  \\
& & $\vec{V}^\top \vec{V} = \vec{I}_K$  \\ \hline
Sparse NMF\tablefootnote{\cite{kim2007SNMF}} & $\frac{1}{2(n-1)} ||\vec{Z} - \vec{U}\vec{V}^\top ||_F^2 + \delta \sum_{i = 1}^n||\vec{u}_i||_2^2 + \lambda \sum_{j = 1}^p ||\vec{v}_j||_1^2$ & $v_{jk} \geq 0, \ \forall j,k$ \\
& & $u_{ik} \geq 0, \ \forall i,k$ \\ \hline
Sparse Orthogonal  & $\frac{1}{2(n-1)} ||\vec{Z} - \vec{U}\vec{V}^\top ||_F^2 + \lambda \sum_{j = 1}^p ||\vec{v}_j||_1$ & $v_{jk} \geq 0, \ \forall j,k$ \\
NMF\tablefootnote{\cite{dai2018}} & & $u_{ik} \geq 0, \ \forall i,k$  \\
& & $\vec{V}^\top \vec{V} = \vec{I}_K$  \\ \hline
Nonnegative Sparse PCA\tablefootnote{\cite{zass2007SNPCA}} & $\frac{1}{2(n-1)}||\vec{X}\vec{V}||_F^2 - \lambda \sum_{j = 1}^p ||\vec{v}_j||_1 - \gamma ||\vec{V}^\top\vec{V} - \vec{I}_K ||_F^2$ & $v_{jk} \geq 0, \ \forall j,k$   \\ \hline
 Clustering around & $\frac{1}{2(n-1)}||\vec{X} - \vec{U}\vec{V}^\top||_F^2 +  \lambda \sum_{j = 1}^p \text{sd}(\vec{x}_j) ||\vec{v}_j||_1$ & $\vec{V} \in \lbrace 0, 1 \rbrace^{p \times K}$ \\
Latent Variables $K + 1$\tablefootnote{\cite{vigneau2016}} & & $\vec{V} \text{ has orthogonal columns}$ \\
& & $\text{var}(\vec{u}_k) = 1$  \\ \hline
Sparse Orthogonal & $\frac{1}{2(n-1)} ||\vec{X} - \vec{U}\vec{V}^\top ||_F^2 + \lambda \sum_{j = 1}^p ||\vec{v}_j||_1$ & $v_{jk} \geq 0, \ \forall j,k$  \\
Semi-NMF\tablefootnote{Method proposed in this paper} &  & $\vec{V} \text{ has orthogonal columns}$  \\
& & $\text{var}(\vec{u}_k) = 1$  \\ 
    \end{tabular}}
    \caption{Summary of NMF methods.}
    \label{tab:S1-NMF_methods}
\end{table}
\newpage

Although it was not originally presented as an NMF method, Clustering around Latent Variables $K + 1$ (CLV $K + 1$, local mode) \citep{vigneau2016} can be rewritten as an NMF problem that is similar to SOS-NMF. In contrast to SOS-NMF, the cluster memberships $v_{jk}$ are constrained to be binary, making them less informative. In addition, the sparsity penalty is weighted by the standard-deviation of the input variables. As a result, cluster assignments only depend on the correlation between the latent variables and the input variables. This means that modifying the scale of the input variables, as is proposed for the first step of VC-PCR (see Section~\ref{sec:S1-step1}), does not affect the relative importance of variables during clustering. Therefore, unlike for SOS-NMF, prior weighting cannot be used to supervise the variable clustering.

\subsection{Coefficient-Grouping Methods}\label{sec:S1-coeff_group}

Most of the supervised variable clustering methods in the literature rely on a regression framework, identifying variable clusters by pushing regression coefficients towards each other in (absolute) value. These \textit{coefficient-grouping methods} (which are listed in Table \ref{tab:sup_clust_methods} for the sake of completeness) generally optimize a least-squares criterion with two penalties: a sparsity penalty, which encourages null coefficients, and a grouping penalty, which encourages coefficients for certain variables to take the same (absolute) value. The objective function for these methods usually takes the form
\begin{equation}
    J(\boldsymbol{b}) = \frac{1}{2}||\vec{y} - \vec{X}\boldsymbol{b}||_2^2 + \text{Penalty}_S(\boldsymbol{b}; \delta) + \text{Penalty}_G(\boldsymbol{b}; \lambda)
\end{equation}
where $\lambda \geq 0$ and $\delta \geq 0$ are hyperparameters, $\text{Penalty}_S(\boldsymbol{b})$ is a sparsity penalty and $\text{Penalty}_G(\boldsymbol{b})$ is a coefficient-grouping penalty.  With the exception of Cluster Elastic Net (CEN) \citep{witten2014}, these methods identify variable clusters after the vector of coefficients $\boldsymbol{b}$ is estimated: all variables with the same estimated coefficient (in absolute or real value, depending on the method) are assigned to the same cluster.

Some of the coefficient-grouping methods rely on prior graphs. If the graph is not known in advance, a fully connected graph containing all $p(p-1)/2$ possible edges may be used. However, this strategy can be inefficient, even for relatively small $p$, as the computational time can increase exponentially with the number of edges \citep{yang2012}. Non-graph-based methods that penalize sums over all pairs of variables also suffer from efficiency problems. Thus, computation time is one of the major challenges for coefficient-grouping methods.

While in the same spirit as other coefficient-grouping methods, Cluster Elastic Net (CEN) \citep{witten2014} approaches the problem of supervised variable clustering in a unique way. The grouping penalty for Cluster Elastic Net (CEN) contains the objective function for K-means, where the points to cluster are the variables weighted by their respective regression coefficients. At each iteration of the CEN algorithm, pairs of variables in the same cluster have their coefficients pushed towards each other in absolute value, especially if they have high absolute correlation. 

\begin{table}[t]
\resizebox{1\textwidth}{!}{%
\begin{tabular}{l|c|l|l|l|l}
\textbf{Method} & \textbf{HPs} & \textbf{Sparsity Term} & \textbf{Grouping Term} & \textbf{$b_j,b_\ell$ grouping iff} & \textbf{Grouping Type} \\ \hline
Flasso \tablefootnote{\cite{tibshirani2005}} & 2 & $\delta\sum_{j = 1}^p |b_j|$ & $\lambda\sum_{j = 2}^p |b_j - b_{j-1}|$ & adjacent & NonAbsVal \\
CLasso \tablefootnote{\cite{she2010}} & 2 & $\delta\sum_{j = 1}^p |b_j|$ & $\lambda \sum_{1 \leq j < \ell \leq p} |b_j - b_\ell|$ &  & NonAbsVal \\
OSCAR \tablefootnote{\cite{bondell2008}} & 2 & $\delta\sum_{j = 1}^p |b_j|$ & $\lambda \sum_{1 \leq j < \ell \leq p} \frac{1}{2}|b_j -   b_\ell| + \frac{1}{2}|b_j +   b_\ell|$ &  & AbsVal \\
PACS \tablefootnote{\cite{sharma2013}} & 2+ & $\lambda \sum_{j = 1}^p \omega_j |b_j |$ & $\lambda \sum_{1\leq j < \ell \leq p} \omega_{j\ell}^{(+)} |b_j -   b_\ell | + \sum_{1\leq j < \ell \leq p} \omega_{j\ell}^{(-)} |b_j +   b_\ell |$ &  & AbsVal \\
WF \tablefootnote{\cite{daye2009}} & 3 & $\lambda \sum_{j = 1}^p |b_j |$ & $\lambda \sum_{1\leq j < \ell \leq p} \omega_{j\ell} (b_j -   \text{sign}(r_{j\ell})b_\ell)^2$ & & AbsVal\\
SSCFS \tablefootnote{\cite{shen2012}} & 3+ & $\delta\sum_{j = 1}^p P_\tau(b_j)$ & $\lambda \sum_{j, \ell \in \mathcal{E}} P_\tau({|b_j - b_\ell|})$ & shared edge & NonAbsVal \\
GFLasso \tablefootnote{\cite{kim2009}} & 2+ & $\delta\sum_{j = 1}^p |b_j|$ & $\lambda \sum_{j, \ell \in \mathcal{E}} |b_j -   \text{sign}(r_{j\ell})b_\ell|$ & shared edge and $r_{j\ell} > 0$ & AbsVal \\
NCFGS \tablefootnote{\cite{yang2012}} & 2+ & $\delta\sum_{j = 1}^p |b_j|$ & $\lambda \sum_{j, \ell \in \mathcal{E}} \bigg|{|b_j|} -   {|b_\ell|}\bigg|$ & shared edge & AbsVal \\
NCTFGS \tablefootnote{\cite{yang2012}} & 3+ & $\delta\sum_{j = 1}^p P_\tau(b_j)$ & $\lambda \sum_{j, \ell \in \mathcal{E}} P_\tau(\bigg|{|b_j|} -   {|b_\ell|}\bigg|)$ & shared edge & AbsVal \\
CEN \tablefootnote{\cite{witten2014}} & 3 & $\delta\sum_{j = 1}^p |b_j|$ & $\lambda \sum_{k = 1}^K \sum_{j \in C_k}\|  \vec{x}_jb_j -   \frac{1}{|C_k|} \sum_{\ell \in C_k}   \vec{x}_\ell   b_\ell \|_2^2$ & $|r_{j\ell}| \rightarrow 1$ and $j,\ell \in C_k$ & AbsVal
\end{tabular}}
\caption{Summary of coefficient-grouping methods. $r_{j\ell} = \text{cor}(\vec{x}_j,\vec{x}_\ell)$; $\mathcal{E}$ is the set of edges in a given graph; $P_\tau(x) = \min(\frac{x}{\tau}, 1)$ for some $\tau > 0$; $\omega_{j}$, $\omega_{j\ell}^{(+)}$ and $\omega_{j\ell}^{(+)}$ are weights that depend on the chosen PACS approach; $\omega_{j\ell} = \frac{|r_{j\ell}|^\gamma}{1-|r_{j\ell}|}$ for some $\gamma > 0$; $C_k$ is the set of variables assigned to cluster~$k$. The grouping type for methods that push coefficients towards each other in absolute value (resp. real value) is ``AbsVal'' (resp. ``NonAbsVal''). For methods involving graphs, the number of hyperparameters (HPs) is presented with a + to indicate that more HPs may be necessary for generating the required graph.}\label{tab:sup_clust_methods}
\end{table}

CEN differs from the other methods with respect to two important aspects. First, the grouping penalty does not require summation over pairs of variables, simplifying the calculations necessary. Second, CEN simultaneously optimizes the regression coefficients and cluster assignments, rather than defining the  clusters after estimating the coefficients. 

Let the variable partition $\mathcal{C} := \lbrace C_1,..., C_K \rbrace$. The objective function for CEN is

\begin{equation}
\label{eq:S1-CEN}
\begin{aligned}
    J(\boldsymbol{b}, \mathcal{C}) = & \frac{1}{2}||\vec{y} - \vec{X}\boldsymbol{b}||_2^2 + \text{Penalty}_S(\boldsymbol{b}; \delta) + \text{Penalty}_G(\boldsymbol{b}, \mathcal{C}, \vec{X}; \lambda)\\
    = & \frac{1}{2}||\vec{y} - \vec{X}\boldsymbol{b}||_2^2 + \delta\sum_{j = 1}^p |b_j| + \lambda \sum_{k = 1}^K \sum_{j \in C_k}\|  \vec{x}_jb_j -   \underbracket{\frac{1}{p_k} \sum_{\ell \in C_k}   \vec{x}_\ell   b_\ell}_{\text{centroid $k$}} \|_2^2, 
\end{aligned}
\end{equation}
where $p_k$ is the number of variables in cluster $k$.

For fixed $\boldsymbol{b}$, the CEN problem reduces to minimizing the grouping penalty function with respect to $\mathcal{C}$. This is equivalent to clustering the coefficient-weighted variables $\vec{x}_1b_1,..., \vec{x}_pb_p$ using K-means. 
As we will demonstrate below, this weighted K-means problem is closely related to our proposed Weighted SOS-NMF problem. Let $\vec{H}$ ($p \times K$) be a binary matrix with orthogonal columns, and let $C_k := \lbrace j \ | \ h_{jk} = 1 \rbrace$. Given a vector $\boldsymbol{b}$ of regression coefficients, the CEN solution for $C_1,...,C_K$ can also be found by estimating $\vec{H}$ and inferring the clusters from it. The solution for $\vec{H}$ is given by

\begin{equation}
\label{eq:S1-CEN_MF}
\begin{aligned}
   \widehat{\vec{H}} = & \argmin_{\vec{H}} \ ||\vec{X}\text{diag}(\boldsymbol{b}) - \vec{X}\text{diag}(\boldsymbol{b})\vec{H}(\vec{H}^\top\vec{H})^{-1}\vec{H}^\top ||_2^2 \\
    & \phantom{\text{s.t. }} \vec{H} \in \lbrace 0, 1 \rbrace^{p \times K} \\ 
    & \phantom{\text{s.t. }} \vec{H} \text{ has orthogonal columns}, \\ 
\end{aligned}
\end{equation}
where $\text{diag}(\boldsymbol{b})$ is a diagonal matrix whose diagonal is composed of the vector $\boldsymbol{b}$ of coefficients.  The CEN solution for $C_1,...,C_K$ can be found by setting $\widehat{C}_k := \lbrace j \ | \ \widehat{h}_{jk} = 1 \rbrace$, $\forall k$. The proof of this proposition is presented in the Supplementary Material (Section~\ref{sec:S1-CEN_proof}).

Clearly, the problem in Eq.~\eqref{eq:S1-CEN_MF} is similar to the Weighted SOS-NMF problem, with a few exceptions: (i) the cluster membership matrix $\vec{H}$ is binary and (ii) the vector $\boldsymbol{b}$ used to weight the variables in $\vec{X}$ is the minimizer of Eq.~\eqref{eq:S1-CEN} for fixed $\mathcal{C}$, not a fixed input. For fixed $\mathcal{C}$ and $b_\ell$, $\ell \neq j$ and $j \in C_k$, the minimizer $\widehat{b}_j$ of Eq.~\eqref{eq:S1-CEN} is

\begin{equation}\label{eq:S1-sol_CEN}
    \widehat{b}_j = \frac{\text{Soft}(\vec{e}_j^\top \vec{x}_j + \frac{\lambda}{p_k}\sum_{\ell \in C_k, \ell \neq j} b_\ell \vec{x}_j^\top\vec{x}_\ell,  \delta)}{\vec{x}_j^\top\vec{x}_j(1 + \frac{\lambda (p_k - 1)}{pk})},
\end{equation}

where $\vec{e}_j = \vec{y} - \sum_{\ell \neq j} \vec{x}_\ell b_\ell$ and $\text{Soft}(z, \delta) = \text{sign}(z)(|z| - \delta)_{+}$. 

As $\lambda$ decreases, the vector $\boldsymbol{b}$ becomes sparser. However, less importance is also given to the variable clustering task, resulting in weaker clustering performance. Therefore, unlike Weighted SOS-NMF, CEN generally achieves better clustering performance for less sparse models, regardless of the true underlying model complexity. This makes it difficult for CEN to simultaneously maximize the quality of variable selection and variable clustering. 

\subsection{Ensemble Methods and SRR}\label{sec:S1-ensemble}

Another approach to the supervised variable clustering problem is to simultaneously estimate $K$ models, one for each cluster, then predict the target based on the average coefficient vector across clusters. Split Regularized Regression (SRR) \citep{christidis2020} takes this ``ensemble-method'' approach. Rather than estimating a single vector of coefficients, SRR estimates multiple sparse coefficient vectors and encourages these vectors to be as diverse as possible. Although SRR does not explicitly search for variable clusters, they may be inferred from the coefficient vectors: a variable belongs to a given cluster if its coefficient in the corresponding coefficient vector is nonzero. This approach can identify overlapping clusters and does not force coefficients of variables in the same cluster to tend towards the same value. 

The objective function for SRR is
\begin{equation}\label{eq:S1-SRR}
\begin{aligned}
      J(\boldsymbol{b}_1, ..., \boldsymbol{b}_K)  =  &\sum_{k = 1}^K \bigg\lbrace \frac{1}{2n}||\vec{y} - \vec{X}\boldsymbol{b}_k||_2^2 + \delta \left[\alpha \sum_{j = 1}^p |b_{jk}| + (1 - \alpha)\sum_{j = 1}^p b_{jk}^2 \right]+ \\
      &+\lambda \sum_{g \neq k} \sum_{j = 1}^p |b_{jk}| |b_{jg}| \bigg\rbrace. \\
\end{aligned}
\end{equation}

The matrix $\vec{B} = \left[\boldsymbol{b}_1 \ ... \ \boldsymbol{b}_K \right]$ is essentially a cluster membership matrix, where variable $j$ belongs to cluster $k$ if $b_{jk} \neq 0$. Variables can belong to multiple clusters, but ``hard'' clusters $C_1, ..., C_K$ can also be defined such that $C_k := \lbrace j \ | \ k = \argmax_{\ell} |b_{j \ell}| \rbrace$. Maximal diversity is achieved when the rows of $\vec{B}$ contain only one nonzero element and thus each variable belongs to a single cluster (i.e.\@ $|b_{jk}| |b_{jg}| = 0 \ \forall j,k,g$).

The final vector of regression coefficients used for prediction is an average across all vectors $\boldsymbol{b}_k$:
\begin{equation}
    \overline{\boldsymbol{b}} = \frac{1}{K} \sum_{k = 1}^K \boldsymbol{b}_k.
\end{equation}

Like many of the coefficient-grouping methods, cluster assignment is more of a side effect than a principal objective of SRR. Moreover, the best solutions in terms of prediction error tend to be complex models with extensive cluster overlap. As a result, variables are assigned to clusters with less certainty, and the clustering performance is less optimal.

\subsection{Two-Step Methods}
\label{sec:S1-two_step}

The methods described in Sections \ref{sec:S1-coeff_group} and \ref{sec:S1-ensemble} are \textit{embedded} methods because model estimation and clustering are performed simultaneously. As a result, the tasks of clustering and modeling are inherently dependent. For two-step methods, these tasks are separated: first, clusters are identified using an unsupervised clustering technique (Step 1), then a model is estimated based on the outputs of the clustering algorithm (Step 2). 

One advantage of two-step methods is their computational efficiency compared to embedded methods. It is also easier to adapt them to different types of prediction tasks, such as regression, classification and survival analysis. Indeed, the outputs of the clustering algorithm can simply be used as inputs to existing prediction algorithms. However, unlike VC-PCR, most two-step methods perform clustering without any supervision by the target. As a result, they cannot distinguish variables that are relevant for prediction from irrelevant ones. This is a significant limitation for these methods, as relevant and irrelevant variables may be mixed into the same cluster, thereby diminishing subsequent prediction performance.

Different unsupervised clustering algorithms have been proposed for the first step of the two-step approach, the most common being K-means (used by e.g.\@ \citet{ma2007}) and hierarchical agglomerative clustering (HAC) (used by e.g.\@ \citet{buhlmann2013,park2007clust}). A variety of different strategies for the second step have also been proposed. These strategies are summarized below. 

For Cluster Representative Lasso (CRL) \citep{buhlmann2013,park2007clust}, Lasso is used to estimate a sparse model in the second step, where the cluster centroids $\widehat{\vec{m}}_1, ..., \widehat{\vec{m}}_K$ from first step are the predictors. The Lasso solution is given by

\begin{equation}
    \argmin_{\vec{a}} L(\widehat{\vec{M}}, \vec{y}; \vec{a}) + \delta ||\vec{a}||_1,
\end{equation}
where $L(\widehat{\vec{M}}, \vec{y}; \vec{a})$ is a loss function, $\vec{a}$ ($K \times 1$) contains the model coefficients and $\delta$ is a nonnegative hyperparameter for the Lasso penalty. 

Performing regression or classification based on the $K$ cluster centroids, rather than the $p$ variables in $\vec{X}$, can be advantageous because $K$ is typically much smaller than $p$. This decreases the dimensionality of the problem, making it less prone to the curse of dimensionality.

For Cluster Group Lasso (CGL) \citep{buhlmann2013}, Group Lasso \citep{yuan2006} is used in the second step. The variables in $\vec{X}$, rather the cluster centroids, are used as predictors, and the clusters $\widehat{C}_1,...,\widehat{C}_K$ are used to define the groups used in the grouping penalty. The Group Lasso solution is given by

\begin{equation}
    \argmin_{\vec{b}} L(\vec{y}, \vec{X}, \vec{b}) + \lambda \sum_{k = 1}^K \sqrt{\sum_{j \in \widehat{C}_k} b_j^2},
\end{equation}
where $L(\vec{y}, \vec{X}, \vec{b})$ is a loss function, $\vec{b}$ ($p \times 1$) contains the model coefficients, $\lambda$ is a nonnegative hyperparameter for the Group Lasso penalty and $\widehat{C}_k$ is the $k^{th}$ cluster identified in the first step. For each cluster $\widehat{C}_k$, the Group Lasso penalty results in either the selection of all of the variables in the cluster (i.e. they all have nonzero coefficients) or none of them (i.e. they all have null coefficients).

An alternative to CRL and CGL is to estimate independent Lasso models for each cluster identified in Step 1. This strategy, which allows for feature selection within clusters, is used for Supervised Group Lasso (SGLasso) \citep{ma2007}. SGLasso also performs a third step where these sparse clusters are used to define the groups in Group Lasso \citep{yuan2006}. The Group Lasso step is necessary for selecting the clusters that are important for predicting the response.

All of the previously mentioned methods are limited by the fact that their clustering step is unsupervised. As a result, clusters may contain both relevant and irrelevant variables, which is undesirable.

\section{Experiments}\label{sec:exp}

Experiments with real and simulated data are performed to compare the performance of the proposed method, VC-PCR, with related methods from the literature. Performance in terms of model error and complexity is evaluated for all methods and datasets. Variable selection and clustering performance is also evaluated for the simulated datasets, for which the true model support and variable clusters are known by design.

\subsection{Methods Compared}

Cluster Elastic Net (CEN) \citep{witten2014} is chosen to represent the coefficient-grouping methods because of its computational efficiency compared to the other methods in this category. Split Regularized Regression (SRR) \citep{christidis2020}, the only ensemble method that can be used to cluster variables, is also included. Of the different two-step approaches, Cluster Representative Lasso (CRL) \citep{buhlmann2013} is tested because it is the most similar to the proposed method. For CRL, three clustering algorithms are tested: hierarchical agglomerative clustering (HAC) based on Ward's criterion (CRL-hclust), HAC based on canonical correlation, as seen in \citet{buhlmann2013} (CRL-hclustCC), and K-means clustering (CRL-kmeans). For VC-PCR, three choices of weight matrices $\vec{W}$ are tested: a diagonal matrix of Lasso regression coefficients (VC-PCR-Lasso), a diagonal matrix of Ridge regression coefficients (VC-PCR-Ridge) and an identity matrix (VC-PCR-Identity). This last choice serves as a benchmark for assessing the added-value of variable-weighting in VC-PCR.

\subsection{Simulated Data}

The simulated datasets vary with respect to three factors: the type of correlation structure among predictors (the configuration), the magnitude of correlation between correlated predictors ($\rho$) and the number of observations ($n$). The simulated data are generated as follows. For each observation $i:1,...,n$, a vector $\vec{x}_i$ is generated from the distribution $N_p(\vec{0}, \boldsymbol{\Sigma})$ and a model error $\varepsilon_i$ is generated from $N(0, \sigma^2_\varepsilon )$, where $\boldsymbol{\Sigma}$ is the covariance matrix for $\vec{X}$ and $\sigma^2_\varepsilon$ is the model noise variance. The response $y_i$ is calculated as $y_i = \vec{x}_i^\top \boldsymbol{b} + \varepsilon_i$, where $\boldsymbol{b}$ is the vector of true model coefficients. The number of observations $n \in \lbrace 25, 50 \rbrace$ and $p$, the number of variables, is fixed to 200.

\subsubsection{True Coefficients}

The vector of true coefficients is fixed for all experiments, meaning that the set of relevant or ``active'' variables $\mathcal{S} := \lbrace j \ | \ b_j \neq 0 \rbrace$ is the same for all datasets. The true vector of coefficients is
\begin{equation}
    \boldsymbol{b} = (\underbracket{1,...,1}_{5}, \underbracket{0,...,0}_{5},\underbracket{-1,...,-1}_{5}, \underbracket{0,...,0}_{5}, \underbracket{1,...,1}_{5}, \underbracket{0,...,0}_{5},\underbracket{-1,...,-1}_{5}, \underbracket{0,...,0}_{5}, \underbracket{0,...,0}_{160})^\top.
\end{equation}
Thus, four blocks of variables are active in the model.

\subsubsection{Covariance Matrix}

Various different covariance matrices $\boldsymbol{\Sigma}$ are used to asses how correlation structure among predictor variables impacts performance. The theoretical variance of all variables is set to one. Active variables are split into $R$ sets $G_1, ..., G_{R}$ such that for each pair of variables  $j,\ell \in \mathcal{S}, j \neq \ell$,

\begin{equation}
\sigma_{j\ell} =
    \begin{cases}
    \rho & \exists r \text{ s.t. } j,\ell \in G_r ; \\
    0 & \exists r \neq r' \text{ s.t. } j \in G_r, \ell \in G_{r'},
    \end{cases}
\end{equation}
where $\rho \in \lbrace 0.3, 0.6 \rbrace$. As a result, variables belonging to different sets $G_r$ and $G_{r'}$ are uncorrelated and variables belonging to the same set have a correlation of $\rho$. 

\begin{figure}[t]
\begin{tabular}{cccc}
& \textbf{Config 1} & \textbf{Config 2} &  \textbf{Config 3}\\ 
{\includegraphics[scale = 0.38,  trim={0 0 13.5cm 0}, clip]{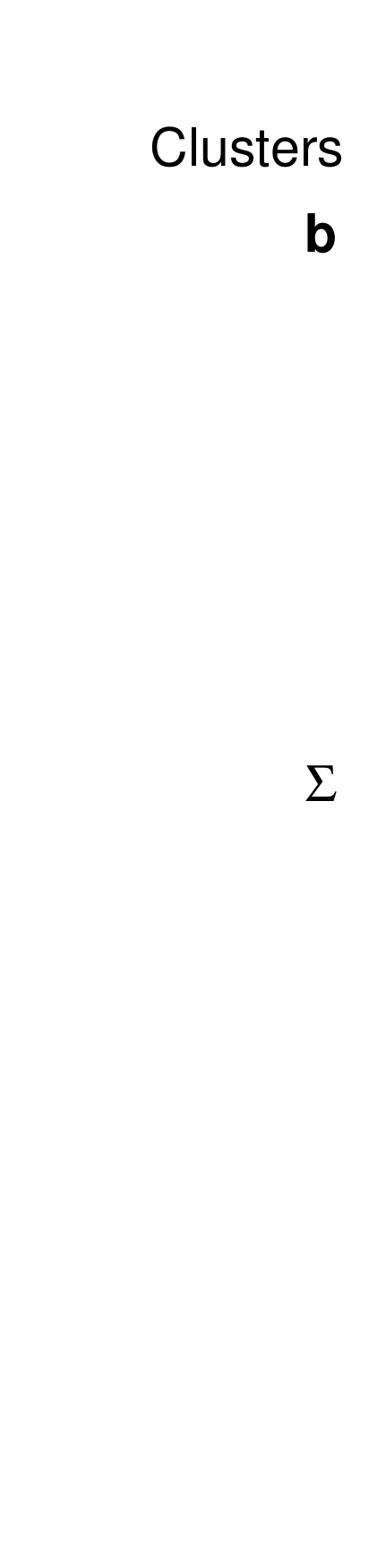}} & {\includegraphics[scale = 0.38,  trim={4.7cm 0 1.8cm 0}, clip]{img/config_1.pdf}} &
{\includegraphics[scale = 0.38,  trim={4.7cm 0 1.8cm 0},clip]{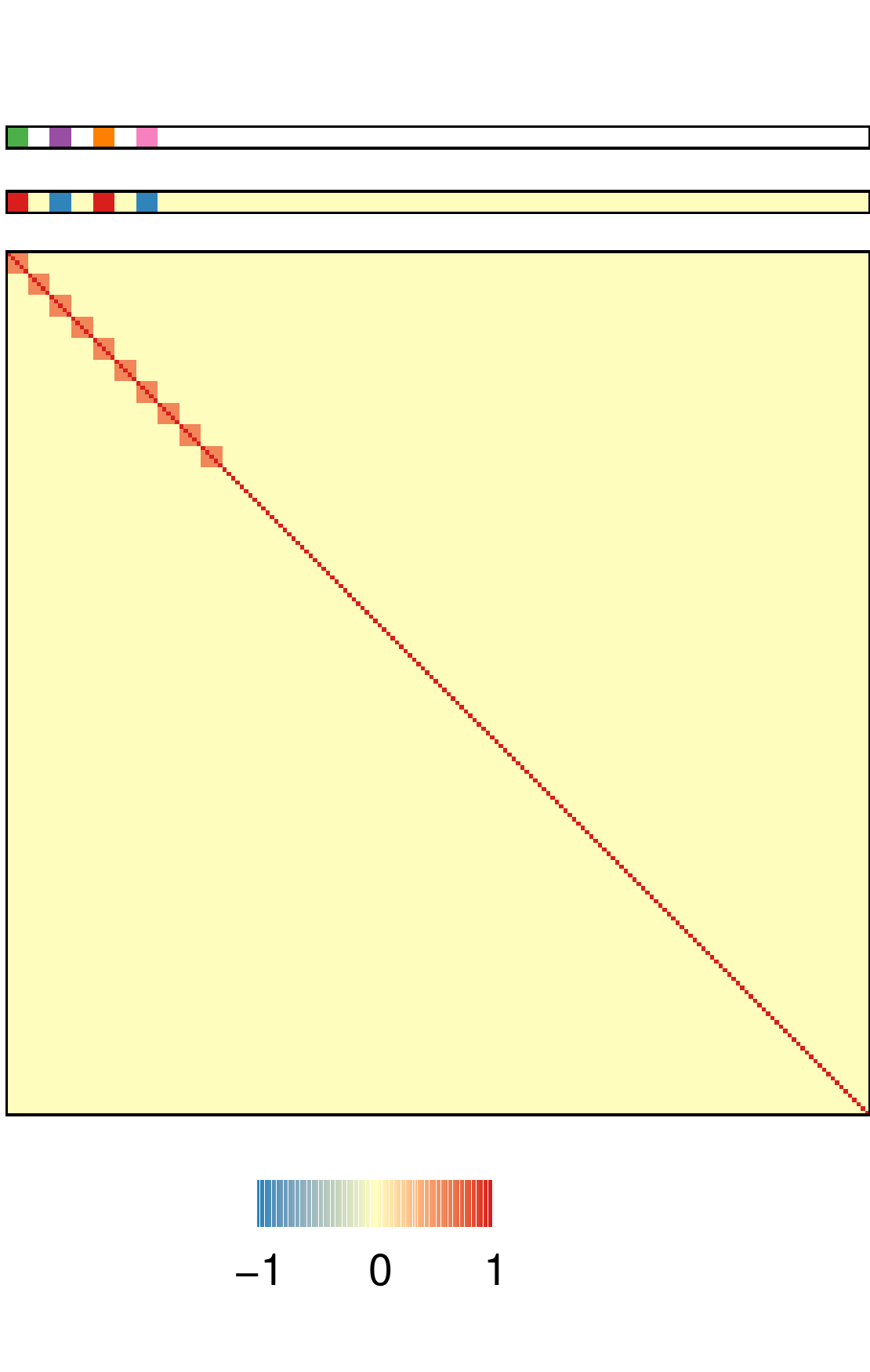}} & {\includegraphics[scale = 0.38,  trim={4.7cm 0 1.8cm 0},clip]{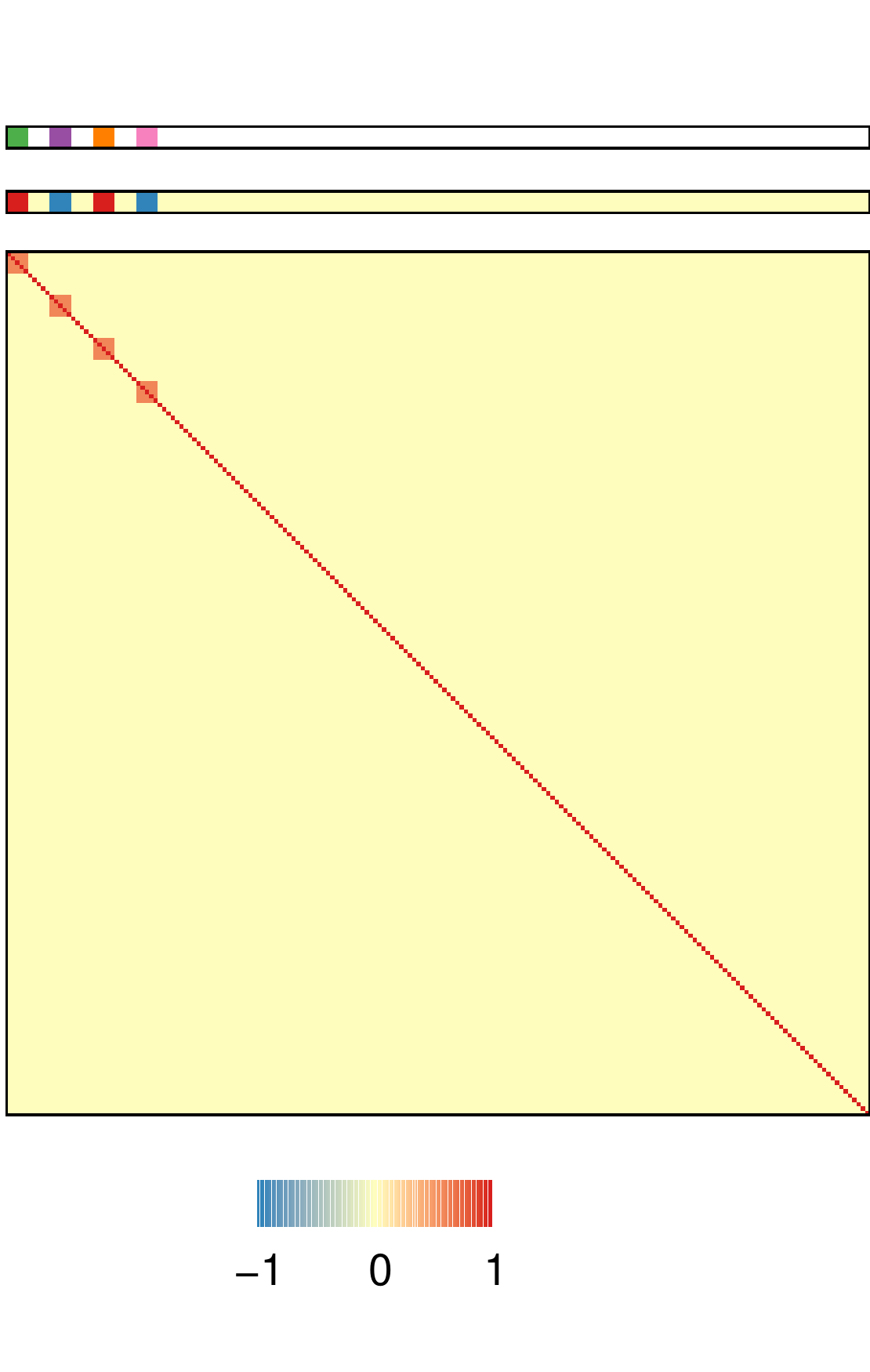}} \\
\end{tabular}
\caption{Experimental configurations for the simulation study. For each configuration, the first and second horizontal bars depict the true cluster assignments and coefficients for the $p = 200$ variables. The square box below is a heatmap of the covariance matrix, $\boldsymbol{\Sigma}$. The colors for the coefficients and covariance heatmap represent continuous values (see the legend), and the colors for cluster assignments are arbitrary, representing categories, i.e.\@ the different clusters $k \in \lbrace 1,...,K \rbrace$.}\label{fig:S1-configs}
\end{figure}

Previous works in the literature have tested different types of correlation structure for inactive variables, but they have not compared these different types in a single study. For inactive variables (i.e.\@ $b_j = 0$), we test three common configurations:

\begin{enumerate}
    \item Some inactive variables are correlated (correlation of $\rho$) with both active variables and other inactive variables (set-up seen in \citet{witten2014})
    \item Some inactive variables are correlated (correlation of $\rho$) with each other but not with active variables (set-up seen in \citet{shen2012})
    \item Inactive variables are uncorrelated with all other variables  (set-up seen in \citet{sharma2013})
\end{enumerate}
These three configurations are summarized in Figure~\ref{fig:S1-configs} for the setting where $\rho = 0.6$.  While the true cluster assignments and coefficients are the same for all configurations, the correlation structure, as seen in the heatmaps of the covariance matrices, varies from one configuration to the next.

\subsubsection{True Clusters}

It is common practice to consider that all inactive variables belong to the same cluster. Moreover, two active variables $j$ and $\ell$ are assumed to be in the same cluster if their absolute correlation is large and their coefficients are similar in absolute value. Therefore, in this experimental set-up, there are $K = R + 1$ true variable clusters $C_k$, one cluster for each set $G_r$ of active variables and one cluster for the set of inactive variables. The true clusters are the same for each configuration, as seen in Figure~\ref{fig:S1-configs}. 

\subsubsection{Variance of the Model Error}

For each configuration, the variance of the model error, $\sigma^2_\varepsilon$, is fixed such that all configurations have the same model signal-to-noise ratio (SNR). The SNR is defined as

\begin{equation}
    SNR = \frac{\boldsymbol{b}^\top \boldsymbol{\Sigma}\boldsymbol{b}}{\sigma^2_\varepsilon}.
\end{equation}

\subsubsection{Summary of Simulation Parameters}

In addition to the true model coefficients and clusters, several parameters are fixed for all simulation settings. These fixed parameters, as well as the three varying parameters, are summarized in Table \ref{tab:sim_params}. A total of $3\times2\times 2 = 12$ different datasets, one for each unique combination of simulation parameters, is generated.

\begin{table}[ht]
    \centering
\begin{tabular}{ c | c | c | l}
\textbf{Type} & \textbf{Parameter} & \textbf{Values} &  \textbf{Meaning}\\
\hline
Varying & Config & $\lbrace 1, 2, 3 \rbrace$ & type of correlation structure \\
Varying & $\rho$ & $\lbrace 0.3, 0.6 \rbrace$ & correlation between correlated variables \\
Varying & $n$ & $\lbrace 25, 50 \rbrace$ & number of observations \\
Fixed & $p$ & 200 & number of variables \\
Fixed & $K$ & 5 & number of true variable clusters \\
Fixed & $SNR$ & 10 & model signal-to-noise ratio \\
\end{tabular}
    \caption{Summary of simulation parameters}
    \label{tab:sim_params}
\end{table}

\subsection{Real Data}

\begin{figure}
    \centering
    \includegraphics[scale = 0.4]{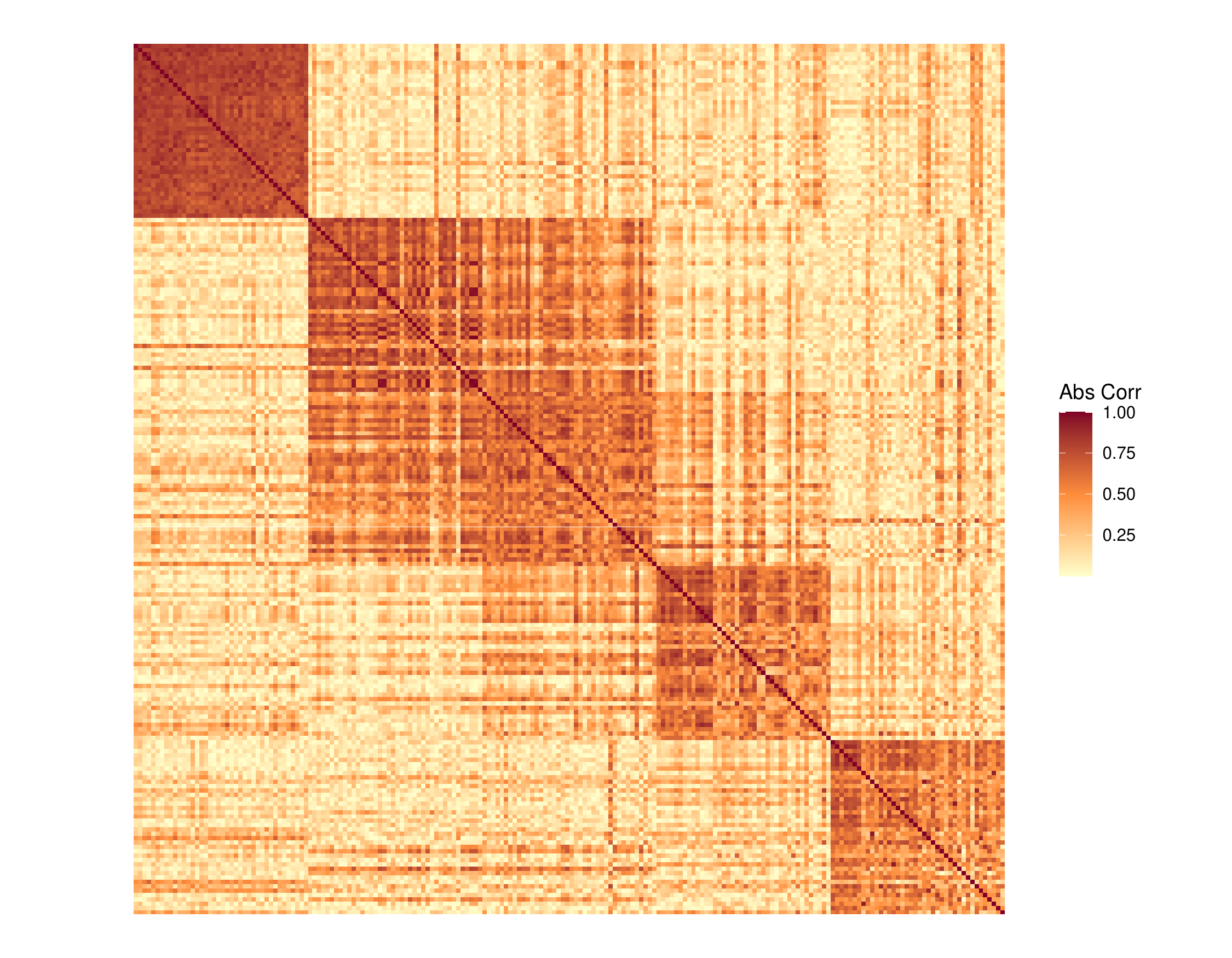}
    \caption{Heatmap of the absolute sample correlations between the variables in the \texttt{Liver} dataset.}
    \label{fig:heatmap_liver}
\end{figure}

We also would like to show that our method can be readily applied to genomics data.
We consider the \texttt{Liver} dataset from \citet{ghazalpour2006}, which was retrieved from \url{https://horvath.genetics.ucla.edu/html/CoexpressionNetwork/Rpackages/WGCNA/Tutorials/}. The \\$3600$ predictor variables in the original dataset are liver gene expression levels for $135$ mice that were subjected to a Western diet over a period of 16 weeks. The response variable is the weight of the mice divided by their length.
Given that this dataset has been analyzed many times already,
we do not add another deep analysis but rather aim at illustrating how our method can benefit gene analyses in general. 
We subsample data as follows:
First, all genes with missing values are excluded, as well as one outlier observation (``$F2\_221$'') and all observations with missing weights or lengths. Among the remaining observations, $n = 50$ are randomly selected for analysis. The normalized predictor matrix containing these observations is denoted $\tilde{\vec{X}}$ and the normalized response vector is denoted $\vec{y}$.
The variables are then further filtered based on Partial Least Squares (PLS) applied to $\tilde{\vec{X}}$ and $\vec{y}$ with the number of components $m = 5$. The resulting loadings matrix for $\tilde{\vec{X}}$, denoted  $\vec{Q}$ ($p \times m)$, is used to assess the potential importance of variables: for each component $r:1,..., m$, the first 40 variables $j$ with the largest absolute loadings, $|q_{jr}|$, are selected as candidate predictors. The variables with the largest loadings in absolute value are the ones that are most important for reconstructing $\tilde{\vec{X}}$, therefore, they contain the most important information from the original dataset. For a given component $r$, the selected variables tend to be highly correlated, whereas for different components, the selected variables tend to be less correlated, although some correlation and selection overlap is observed. In total, $p = 195$ unique predictors are selected for the final matrix $\vec{X}$. The sample correlation matrix for $\vec{X}$ (presented in absolute value) is shown in Figure~\ref{fig:heatmap_liver}.

\subsection{Evaluation Criteria}

The performance of each method is evaluated with respect to four different characteristics: the model error, the model complexity, the recovery of the model support and the recovery of the underlying clusters. The model error is quantified with the Mean Squared Error of Prediction ($\operatorname{MSEP}$). Given a vector $\widehat{\vec{y}}$ ($n_{test} \times 1$) of predicted response values,

\begin{equation}
    \operatorname{MSEP} = \frac{1}{n_{test}} \sum_{i = 1}^{n_{test}} (y_i - \widehat{y}_i)^2.
\end{equation}

The model complexity is quantified as the number $s$ of active variables in the model. For methods that estimate one coefficient $\widehat{b}_j$ per variable $j$ (e.g.\@ CEN, SRR, etc.), 
\begin{equation}\label{eq:s_b}
    s = ||\widehat{\boldsymbol{b}}||_0 = \sum_{j = 1}^p \mathbb{I}[\widehat{b}_j \neq 0 ].
\end{equation}
For methods that estimate one coefficient $\widehat{a}_k$ per cluster $k$ (e.g.\@ CRL, etc.),
\begin{equation}\label{eq:s_a}
    s = \sum_{k = 1}^K  \widehat{p}_k \mathbb{I}[\widehat{a}_k \neq 0 ],
\end{equation}
where $\widehat{p}_k$ is the number of variables in $\widehat{C}_k$.

The Matthews Correlation Coefficient ($\operatorname{MCC}$) \citep{matthews1975} is used to quantify the recovery of the true model support and underlying clusters. Compared to other existing criteria for evaluating binary classification, $\operatorname{MCC}$ is considered to be less biased and is more robust to class imbalance \citep{chicco2017,chicco2020,chicco2021}. Given the number of true positives (TP), true negatives (TN), false positives (FP) and false negatives (FN), 
\begin{equation}
    \operatorname{MCC} = \frac{(TP)(TN) - (FP)(FN)}{\sqrt{(TP + FP)(TP + FN)(TN + FP)(TN + FN)}} \in \left[-1, 1\right],
\end{equation}
where a score of 1 represents perfect classification and a score of 0 is equivalent to random classification.  

When evaluating model support recovery, variables with nonzero coefficients are assigned to the ``positive'' class and variables with null coefficients are assigned to the ``negative class.'' Cluster recovery is evaluated for all unique pairs of variables. Pairs of variables assigned to the same cluster are in the ``positive class'' and all other pairs of variables are in the ``negative'' class.

\subsection{Hyperparameter Selection}\label{sec:S1-hp_tuning}

The methods compared in this study depend on the choice of several hyperparameters: $\delta$, $\lambda$ and $K$ for VC-PCR and the embedded methods and $\delta$ and $K$ for the CRL methods. For SRR, the fourth hyperparameter $\alpha$ is fixed to 1, resulting in a Lasso penalty (see Eq.~\eqref{eq:S1-SRR}). For the hyperparameters $\delta$ and $\lambda$, 10 values are tested for each method. For the simulated dataset, three values of $K$ are tested: $\lbrace 4, 5, 6 \rbrace$ (the true value is 5). For the real dataset, five values of $K$ are tested: $\lbrace 10, 20, 40, 50, 60 \rbrace$. Some methods, such as CRL-kmeans, CEN and VC-PCR, require an initial set of clusters for each value of $K$. Five random initializations are tested for these methods, and the resulting performance is subsequently averaged (see Section~\ref{sec:S1-results}).

Nested $k$-fold cross-validation is used to select hyperparameters and evaluate average out-of-sample performance. This procedure involves $k$-fold cross-validation at two levels: an inner and an outer loop. The dataset is split into $10$ outer folds. Then, for each iteration $m$ of the outer loop, the samples in outer fold $m$ are set aside and $5$-fold cross-validation (the nested inner loop) is performed on the remaining samples. The latter samples are normalized before analysis: the target and predictors are centered and scaled to unit variance. The hyperparameters yielding the model with the smallest average $\operatorname{MSEP}$ across the inner folds are then selected and a final model is trained using all samples in the inner folds, which are also normalized. The performance of this model is evaluated using the samples in outer fold $m$, which were not used during model selection or estimation. These samples are normalized using the same parameters used to normalize the samples that trained the model. After iterating through all of the outer folds, the average out-of-sample performance is then calculated. 
\subsection{Numerical Implementation}

The experiments were programmed in R\footnote{Code available at \url{https://anonymous.4open.science/r/VC-PCR-Experiments-F994}}. Weighted SOS-NMF, VC-PCR and SRR were implemented using in-house code. CEN was performed using code provided by the authors. Lasso and Ridge regression were performed using the function \textit{glmnet} from the \texttt{glmnet} package. For the CRL methods, Ward's hierarchical clustering and K-means clustering were performed using the functions \textit{hclust} and \textit{kmeans} from the \texttt{stats} package. Hierarchical clustering based on canonical correlation was performed using the function \textit{hclustvar} from the \texttt{ClustOfVar} package.

Of the methods compared, Cluster Representative Lasso (CRL) with K-means clustering (CRL-kmeans) and hierarchical clustering based on Ward's criterion (CRL-hclustCor) were the fastest methods (faster than VC-PCR by a factor of 16 and 7 respectively). This is likely due to the fact that the \textit{kmeans} function from the \texttt{stats} package is partly coded in C and the \textit{hclust} function is partly coded in Fortran. Compared to VC-PCR, Cluster Elastic Net (CEN) was 4 times slower, CRL with hierarchical clustering based on canonical correlation (CRL-hclustCC) was 7 times slower and Split Regularized Regression (SRR) was several 100 times slower.

\subsection{Results}\label{sec:S1-results}

This section presents the results for the simulated data and the real data example. For each dataset and method, all but one of the hyperparameters is tuned using the cross-validation strategy explained in Section~\ref{sec:S1-hp_tuning}, and the average out-of-sample performance is calculated for each value of the remaining hyperparameter. The latter hyperparameter is the one that most impacts model complexity: $\delta$ for the CRL and embedded methods and $\lambda$ for the VC-PCR methods. For methods requiring cluster initializations, performance is further averaged over all random initializations.

\subsubsection{Simulated Data}

\begin{figure}[hp]
    \centering
    \includegraphics[scale = 0.45, trim = {0cm 4cm 0cm 0.25cm}, clip]{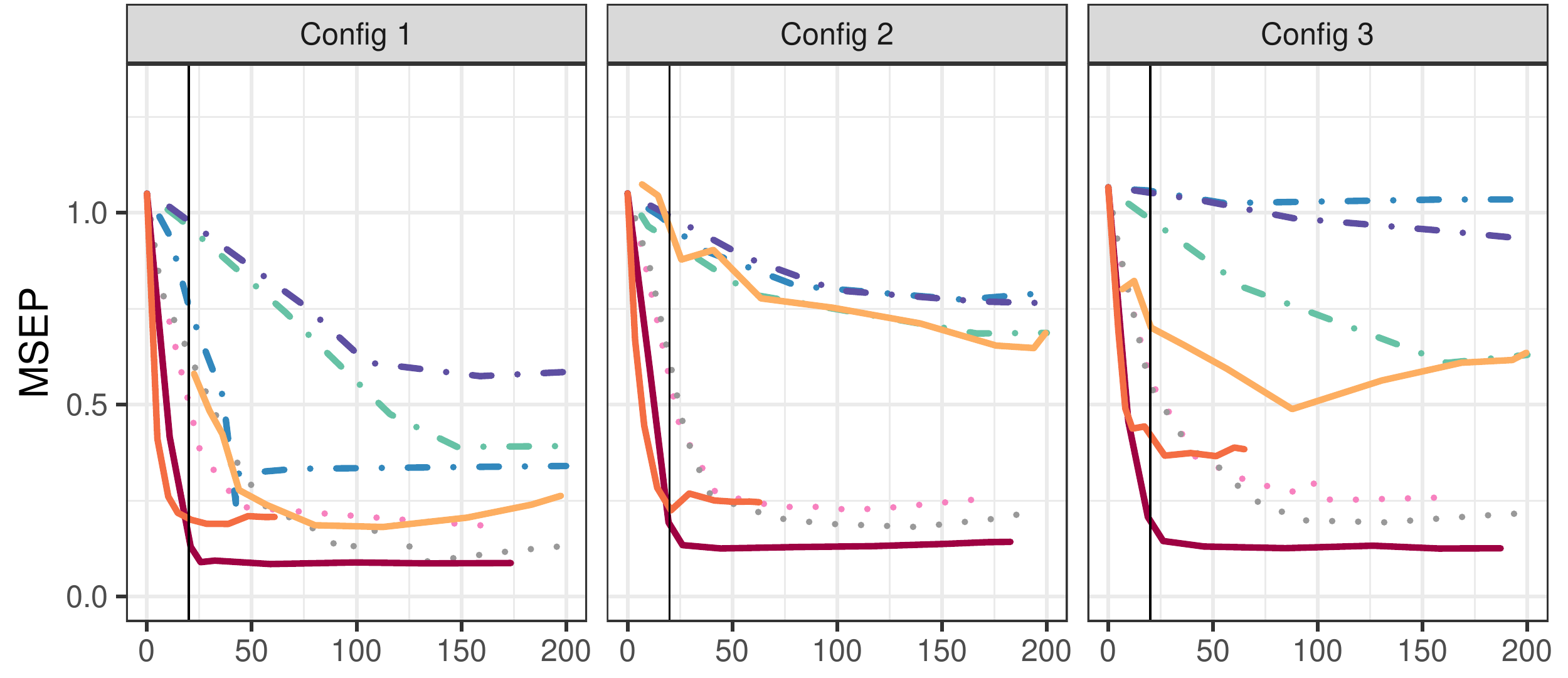}
    
    \includegraphics[scale = 0.45,trim = {0cm 4cm 0cm 0cm}, clip]{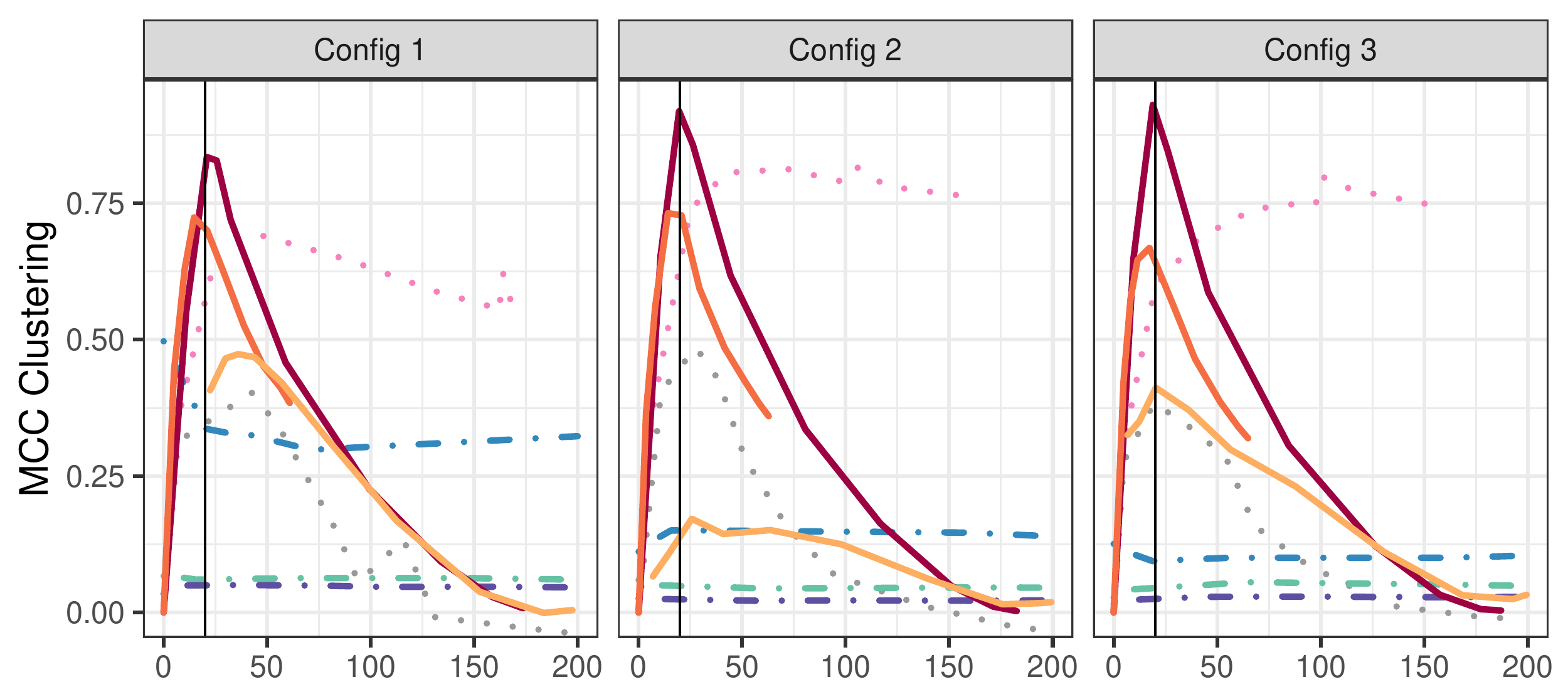}
    
    \includegraphics[scale = 0.45,trim = {0cm 0.5cm 0cm 0cm}, clip]{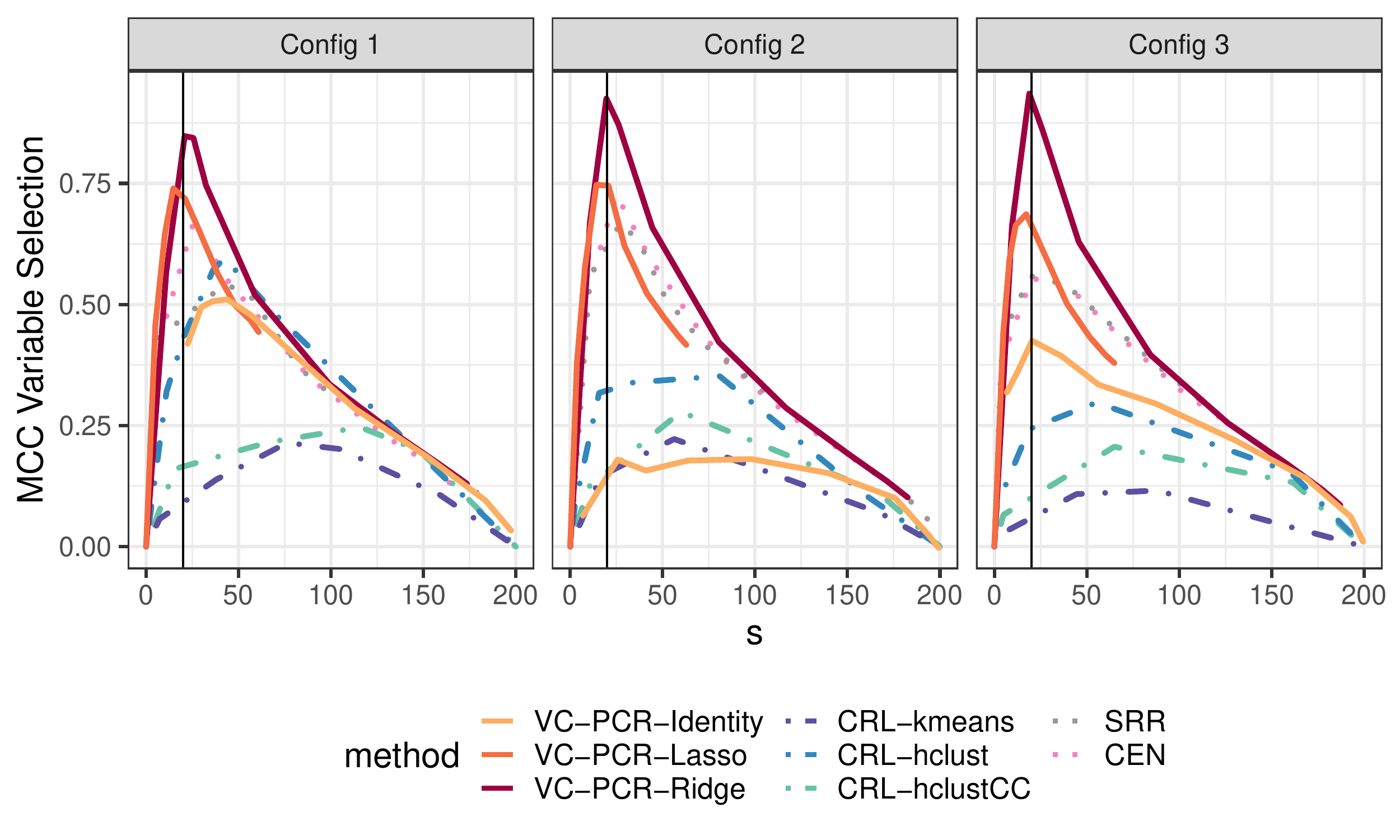}
    \caption{{\small Average out-of-sample performance for $n = 50$ and $\rho = 0.6$, plotted with respect to the average model size $s$. Each point in the plotted lines corresponds to a value of the hyperparameter (HP) $\delta$ or $\lambda$; all other HPs were tuned using $k$-fold cross-validation. The vertical line indicates the true number of active variables, $20$.}}
    \label{fig:S1-perf}
\end{figure}

For each data configuration, sample size $n$ and correlation level $\rho$, the model error, cluster recovery and support recovery are assessed for models with different complexities $s$ (see Eq.~\eqref{eq:s_b} and Eq.~\eqref{eq:s_a}). Figure~\ref{fig:S1-perf} depicts the performance results for the setting where $n = 50$ and $\rho = 0.6$. The average out-of-sample performance is plotted with respect to $s$, where each point in the lines corresponds to a different value of the fixed hyperparameter ($\delta$ or $\lambda$). The vertical black line corresponds to the true number of active variables, $20$. For VC-PCR-Lasso, the plotted line does not cover the entire domain of model sizes $s$. This is because VC-PCR-Lasso depends on two sparsity parameters: $\delta$ for the Lasso regression used to generate the weight matrix for Weighted SOS-NMF and $\lambda$ for the sparsity penalty in Weighted SOS-NMF. For the plots depicted in Figure~\ref{fig:S1-perf}, only one of these hyperparameters is fixed, while the other is tuned using cross-validation. Clearly, hyperparameters yielding larger models are never selected.

For the setting where $n = 50$ and $\rho = 0.6$ (Figure~\ref{fig:S1-perf}), VC-PCR-Ridge outperforms the other methods tested for all of the configurations and performance criteria. For VC-PCR-Ridge, the best solutions in terms of clustering and support recovery are models with a complexity $s$ close to the true number of active variables. Moreover, unlike the other two-step methods (i.e. CRL), the model error for VC-PCR-Ridge is robust to the type of configuration. VC-PCR-Lasso demonstrates a similar level of robustness to configuration type, but its performance is slightly inferior to VC-PCR-Ridge. In contrast to these two methods, the model error for VC-PCR-Identity greatly depends on the configuration type. In addition, the clustering and support recovery are much worse. This demonstrates the added-value of using non-uniform weights in Weighted SOS-NMF, the first step of VC-PCR.

Results for the other simulation settings can be found in the Supplementary Material (Section~\ref{sec:add_sim_results}). The prediction, variable selection and clustering performance of VC-PCR generally improves as the sample size $n$ or the correlation $\rho$ between correlated variables increases. VC-PCR-Ridge achieves the best variable selection and clustering performance for all configurations, sample sizes and correlation levels tested. This top performance occurs for models with a complexity close to the true complexity, $s = 20$ variables. The prediction performance of VC-PCR is most impacted by the correlation $\rho$ between variables, with weaker performance occurring when there is weak correlation between variables (i.e. weak clustering structure).  If $\rho$ is sufficiently high (0.6 in this study), VC-PCR-Ridge has better prediction performance than other methods for all sample sizes, configurations and model complexities tested. This suggests that VC-PCR could be used to test hypotheses about cluster structure in the data: if prediction error is not improved using VC-PCR, then the variables analyzed may not contain predictive cluster structure.

\subsubsection{Real Data}

\begin{figure}
    \centering
    \includegraphics[scale = 0.5]{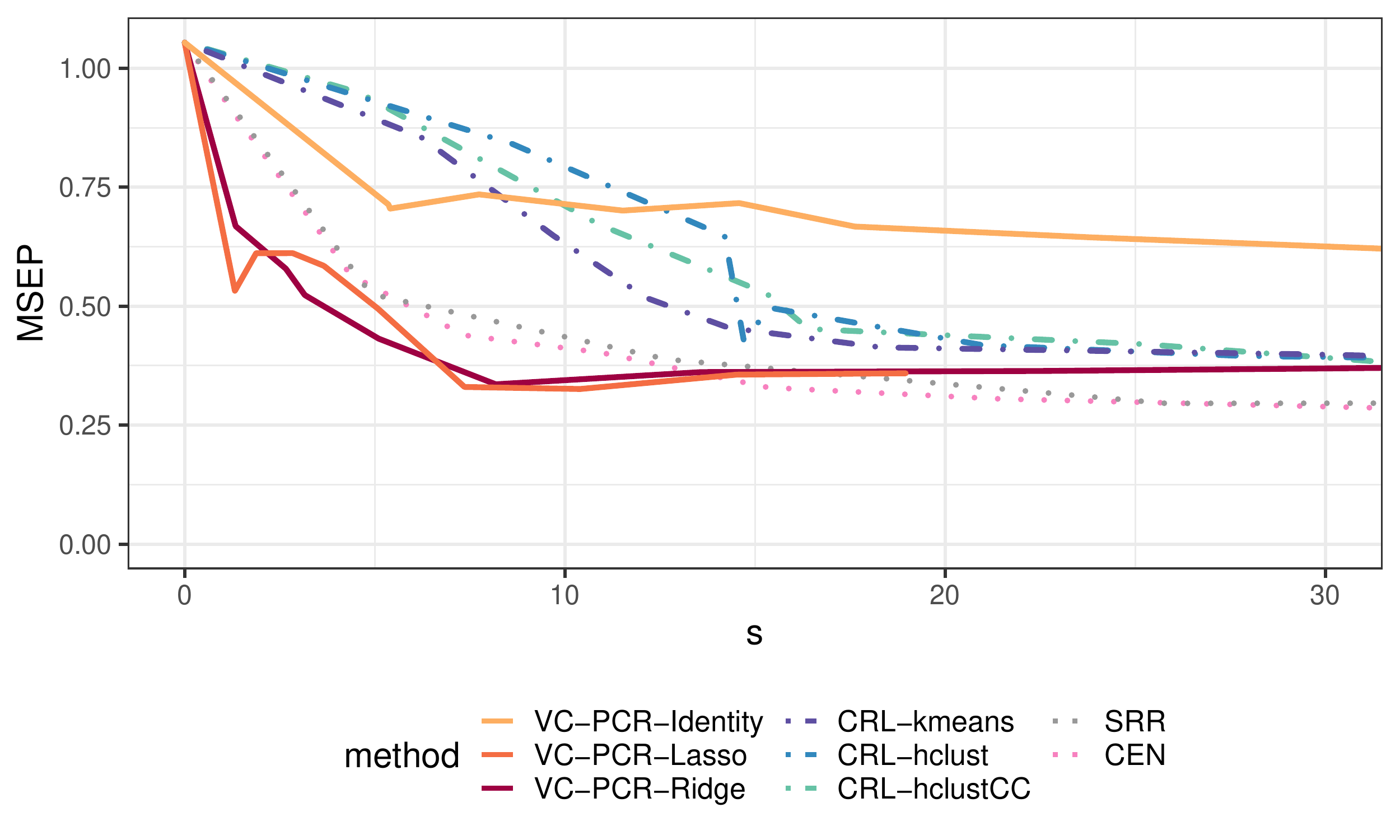}
    \caption{{\small Average out-of-sample $\operatorname{MSEP}$ for the \texttt{Liver} dataset, plotted with respect to the average model size $s$. Each point in the plotted lines represents a value of the hyperparameter (HP) $\delta$ or $\lambda$; all other HPs were tuned using $k$-fold cross-validation.}}
    \label{fig:liver_results}
\end{figure}

For the \texttt{Liver} dataset, the true model support and variable clusters are unknown, so it is not possible to calculate the $\operatorname{MCC}$ for variable selection and clustering. Figure~\ref{fig:liver_results} depicts the average out-of-sample prediction error ($\operatorname{MSEP}$) plotted with respect to the average model complexity ($s$). As with the simulated data, VC-PCR-Ridge has the best prediction performance for sparse models (about 20 or fewer variables).

\section{Discussion and Conclusions}\label{sec:concl}

This paper presented a two-step method for prediction and supervised variable clustering, Variable Cluster Principal Component Regression (VC-PCR). In the first step, VC-PCR identifies variable clusters and performs variable selection via Weighted SOS-NMF. The weights used for Weighted SOS-NMF \textemdash which encode the prior relationship between the predictors and the response \textemdash make it possible to supervise both the clustering and the variable selection. In the second step, the cluster membership degrees from the first step are used to generate latent variables, and a linear model is estimated to relate the latent variables to the response.  

Unlike other two-step methods, the clustering step in VC-PCR is supervised. As a result, the identified clusters tend to be more homogeneous: active and inactive variables are less likely to be mixed together. Compared to embedded methods, VC-PCR is more computationally efficient, and it can be easily applied to a variety of tasks (regression, classification, survival analysis, etc.), not just regression. 

Numerical experiments using real and simulated data were performed to compare VC-PCR to competitors from the literature. For all data configurations, VC-PCR with Ridge regression weights (VC-PCR-Ridge) yielded the best variable selection and clustering performance. For datasets with predictive cluster structure, VC-PCR also achieved superior prediction performance, especially for sparse models. Unlike related methods from the literature, VC-PCR is able to simultaneously improve prediction, variable selection and clustering performance.


\newpage
\bigskip
\begin{center}
{\large\bf SUPPLEMENTARY MATERIAL}
\end{center}

\appendix

\label{sec:append}

\section{Optimization of SOS-NMF}
\label{sec:opt_SOS-NMF}

The solution to the SOS-NMF problem was presented in Section~\ref{sec:opt_SOS-NMF_intext}. This section presents detailed derivations of this solution.

\subsection{Solution for $\vec{U}$}

For fixed $\vec{V}$, the objective function to be minimized is

\begin{equation}
    J(\vec{U}) = \frac{1}{2(n-1)}||\vec{X} - \vec{U}\vec{V}^\top ||_F^2.
\end{equation}
This is a least squares problem, and the solution for an unconstrained $\vec{U}$ is $\tilde{\vec{U}} = \vec{X}\vec{V}(\vec{V}^\top\vec{V})^{-1}$. The unit-variance constraint on the latent variables $\vec{u}_k$ can be accounted for by scaling the vectors $\tilde{\vec{u}}_k$, as explained in Section~\ref{sec:opt_SOS-NMF_intext}.

\subsection{Solution for $\vec{V}$}

Let $C_k:= \lbrace j \ | \ v_{jk} > 0 \rbrace$. For fixed $\vec{U}$, the objective function from Eq.~\eqref{eq:S1-SOS-NMF} becomes

\begin{equation}\label{eq:S1-JV-SOS}
\begin{aligned}
      J(\vec{V}) = & \frac{1}{2(n-1)}||\vec{X} - \vec{U}\vec{V}^\top ||_F^2 + \lambda \sum_{j = 1}^p ||\vec{v}_j||_1 \\
      = & \frac{1}{2(n-1)} \sum_{j = 1}^p \bigg\lbrace c_1 -2\vec{v}_j^\top \vec{U}^\top \vec{x}_j  + \vec{v}_j^\top \vec{U}^\top\vec{U}\vec{v}_j \bigg\rbrace + \lambda \sum_{j = 1}^p||\vec{v}_j||_1 \\
      = & c_2 + \sum_{k = 1}^K \sum_{j \in C_k} \bigg\lbrace \frac{1}{2(n-1)} \bigg(- 2v_{jk} \vec{u}_k^\top \vec{x}_j  + v_{jk}^2 \underbracket{\vec{u}_k^\top\vec{u}_k}_{n-1}\bigg) + \lambda |v_{jk}| \bigg\rbrace \\
      = & c_2 + \sum_{k = 1}^K \sum_{j \in C_k} \bigg\lbrace - \frac{1}{n-1}v_{jk} \vec{u}_k^\top \vec{x}_j  + \frac{1}{2} v_{jk}^2 + \lambda |v_{jk}| \bigg\rbrace,
\end{aligned}
\end{equation}
where $c_1$ and $c_2$ are constants that do not depend on $\vec{V}$.

Suppose that, for a given variable $j$, $j \in C_k$ and thus $v_{jk} > 0$. The solution $\widehat{v}_{jk}$ is found by deriving Eq.~\eqref{eq:S1-JV-SOS} with respect to $v_{jk}$ and setting it to zero.

\begin{equation}
    \frac{\partial J(\vec{V})}{\partial v_{jk}} =   -\frac{1}{n-1} \vec{u}_k^\top \vec{x}_j  + v_{jk} + \lambda = 0.
\end{equation}
Thus, if $j \in C_k$,

\begin{equation}\label{eq:S1-vjk-SOS}
\begin{aligned}
     \widehat{v}_{jk} & = \frac{1}{n-1} \vec{u}_k^\top \vec{x}_j  - \lambda \\
     & = \text{cov}(\vec{u}_k, \vec{x}_j) - \lambda \\
     & = \underbracket{\text{sd}(\vec{u}_k)}_{1}\underbracket{\text{sd}(\vec{x}_j )}_{1}\text{cor}(\vec{u}_k, \vec{x}_j) - \lambda \\
     & = \text{cor}(\vec{u}_k, \vec{x}_j) - \lambda. \\
\end{aligned}
\end{equation}

Thanks to Eq.~\eqref{eq:S1-vjk-SOS}, we can deduce that $\text{cor}(\vec{u}_k, \vec{x}_j) > \lambda$ is a necessary condition for variable $j$ to belong to cluster $k$ (i.e.\@ $v_{jk} > 0$). In addition to satisfying this condition, another necessary condition is that

\begin{equation}
    k = \argmax_{\ell} \text{cor}(\vec{u}_\ell, \vec{x}_j).
\end{equation}
Indeed, thanks to the orthogonality and nonnegativity constraints, at most one element of $\vec{v}_j$ may be greater than zero. This means that, for a given variable $j$, the sum over $k$ in Eq.~\eqref{eq:S1-JV-SOS} is only active for one value $k$.  For $v_{jk} > 0$, the objective function evaluated at the solution $\widehat{v}_{jk}$ is
\begin{equation}\label{eq:S1-JVjk-SOS}
\begin{aligned}
 J(\widehat{v}_{jk}) & =  - \frac{1}{n-1}\widehat{v}_{jk} \vec{u}_k^\top \vec{x}_j  + \frac{1}{2} \widehat{v}_{jk}^2 + \lambda \widehat{v}_{jk}\\
 & = -\widehat{v}_{jk}\underbracket{\text{cor}(\vec{u}_k, \vec{x}_j)}_{\widehat{v}_{jk} + \lambda} + \frac{1}{2} \widehat{v}_{jk}^2 + \lambda \widehat{v}_{jk} \\
  & = - \frac{1}{2} \widehat{v}_{jk}^2  \\
\end{aligned}
\end{equation}
For a fixed variable $j$, the index $k$ with the smallest $J(\widehat{v}_{jk})$ is the one for which $\text{cor}(\vec{u}_k, \vec{x}_j)$ is the largest. 

\section{Other Related NMF Methods}
\label{sec:otherNMF}

As mentioned in Section~\ref{sec:S1-NMF}, certain other NMF methods are related to SOS-NMF. These methods are compared below.

Semi-NMF \citep{ding2008} is similar to SOS-NMF in that only $\vec{V}$ is constrained to be nonnegative (hence the name ``semi''-NMF). However, in contrast to SOS-NMF, Semi-NMF does not constrain the columns of $\vec{V}$ to be orthogonal or impose any sparsity penalties. As a result, for a given variable $j$, all of the elements in $\vec{v}_j$ may be nonzero, meaning that variable $j$ is essentially assigned to all of the clusters. Thus, Semi-NMF is not a candidate method for variable clustering.

As with SOS-NMF, one-sided Orthogonal NMF \citep{ding2006} constrains the columns of $\vec{V}$ to be nonnegative and orthogonal. These two constraints make Orthogonal NMF a candidate method for variable clustering. However, Orthogonal NMF requires the input of a nonnegative matrix $\vec{Z}$. Moreover, no sparsity penalty is applied to $\vec{V}$, making the method less robust to noise variables.

The Sparse NMF method proposed in \citet{kim2007SNMF} (``SNMF/R''), which constrains $\vec{V}$ to be sparse, is another method that bears similarities to SOS-NMF. However, this method does not constrain the columns of $\vec{V}$ to be orthogonal. As a result, each vector $\vec{v}_j$ may contain more than one nonzero element, meaning that each variable $j$ may be assigned to multiple clusters. While the sparsity penalty for the vectors $\vec{v}_j$ can encourage some elements of $\vec{v}_j$ to be zero, in practice, the clustering of variables can be unsatisfactory (see \citet{marion2020}).

\section{Orthonormal Nonnegative Sparse PCA}
\label{sec:S1-orthoNSPCA_proof}

As mentioned in Section~\ref{sec:S1-NMF}, Nonnegative Sparse PCA with $\gamma = \infty$ only differs from SOS-NMF with respect to the scaling constraints on $\vec{U}$ and $\vec{V}$: SOS-NMF constrains the scale of $\vec{U}$ whereas NSPCA constrains the scale of $\vec{V}$. 

\begin{proof}
For $\gamma = \infty$, the NSPCA problem can be rewritten as

\begin{equation}\label{eq:S1-NSPCA_orthog}
\begin{aligned}
    \argmax_{\vec{V}} \ &   \frac{1}{2(n-1)}||\vec{X}\vec{V}(\vec{V}^\top \vec{V})^{-1/2}||_F^2 - \lambda \sum_{j = 1}^p ||\vec{v}_j||_1\\
    & \text{s.t. } v_{jk} \geq 0, \ \forall j,k\\
    & \phantom{\text{s.t. }} \vec{V}^\top\vec{V} = \vec{I}_K,\\
\end{aligned}
\end{equation}
where $||\vec{X}\vec{V}(\vec{V}^\top \vec{V})^{-1/2}||_F^2 = ||\vec{X}\vec{V}||_F^2$ because $\vec{V}^\top\vec{V} = \vec{I}_K$. This can also be rewritten as a minimization problem similar to the one in Eq.~\eqref{eq:S1-JV-SOS}. Note that

\begin{equation}
    \begin{aligned}
||\vec{X} - \vec{X}\vec{V}(\vec{V}^\top \vec{V})^{-1}\vec{V}^\top ||_F^2 = &  
\text{Tr}\bigg(\vec{X}\vec{X}^\top - \vec{X}\vec{V}(\vec{V}^\top \vec{V})^{-1}\vec{V}^\top\vec{X} \bigg)\\
= & ||\vec{X}||_F^2 - ||\vec{X}\vec{V}(\vec{V}^\top \vec{V})^{-1/2}||_F^2.
    \end{aligned}
\end{equation}
Therefore, the solution represented in Eq.~\eqref{eq:S1-NSPCA_orthog} is the same as the solution
\begin{equation}\label{eq:S1-NSPCA_orthog_VV}
    \begin{aligned}
\argmin_{\vec{V}} \ &   \frac{1}{2(n-1)}||\vec{X} - \vec{X}\vec{V}(\vec{V}^\top \vec{V})^{-1}\vec{V}^\top ||_F^2 + \lambda \sum_{j = 1}^p ||\vec{v}_j||_1\\  
 & \text{s.t. } v_{jk} \geq 0, \ \forall j,k\\
    & \phantom{\text{s.t. }} \vec{V}^\top\vec{V} = \vec{I}_K.\\
    \end{aligned}
\end{equation}

Eq.~\eqref{eq:S1-NSPCA_orthog_VV} can also be rewritten as a matrix factorization problem as follows:
\begin{equation}\label{eq:S1-NSPCA_orthog_MF}
    \begin{aligned}
\argmin_{\vec{U}, \vec{V}} \ & \frac{1}{2(n-1)} ||\vec{X} - \vec{U}\vec{V}^\top ||_F^2 + \lambda \sum_{j = 1}^p ||\vec{v}_j||_1\\  
 & \text{s.t. } v_{jk} \geq 0, \ \forall j,k\\
    & \phantom{\text{s.t. }} \vec{V}^\top\vec{V} = \vec{I}_K,\\
    \end{aligned}
\end{equation}
where $\vec{U}$ ($n \times K$) is a matrix of latent variables. For fixed $\vec{V}$, the solution $\widehat{\vec{U}} = \vec{X}\vec{V}(\vec{V}^\top\vec{V})^{-1}$. Replacing the $\vec{U}$ in Eq.~\eqref{eq:S1-NSPCA_orthog_MF} with $\widehat{\vec{U}}$ results in the optimization problem represented in Eq.~\eqref{eq:S1-NSPCA_orthog_VV}, as

\begin{equation}
    ||\vec{X} - \widehat{\vec{U}}\vec{V}^\top ||_F^2 = ||\vec{X} - \vec{X}\vec{V}(\vec{V}^\top\vec{V})^{-1}\vec{V}^\top ||_F^2. 
\end{equation}

Therefore the solution for $\vec{V}$ in Eq.~\eqref{eq:S1-NSPCA_orthog_MF} is indeed the same as the solution in Eq.~\eqref{eq:S1-NSPCA_orthog_VV}. Clearly, this problem is equivalent to the SOS-NMF problem except for the fact that the scale of the columns of $\vec{V}$, rather than $\vec{U}$, is fixed. 
\end{proof}

\section{Cluster Elastic Net Reformulation}
\label{sec:S1-CEN_proof}

As mentioned in Section~\ref{sec:S1-coeff_group}, the CEN solution for $C_1,...,C_K$ can also be found by estimating a binary matrix $\vec{H}$ ($p \times K$) and inferring clusters from it. For a fixed vector $\boldsymbol{b}$ of regression coefficients, the solution for $\vec{H}$ is given by

\begin{equation}
\label{eq:S1-CEN_MF2}
\begin{aligned}
   \widehat{\vec{H}} = & \argmin_{\vec{H}} \ ||\vec{X}\text{diag}(\boldsymbol{b}) - \vec{X}\text{diag}(\boldsymbol{b})\vec{H}(\vec{H}^\top\vec{H})^{-1}\vec{H}^\top ||_2^2 \\
    & \phantom{\text{s.t. }} \vec{H} \in \lbrace 0, 1 \rbrace^{p \times K} \\ 
    & \phantom{\text{s.t. }} \vec{H} \text{ has orthogonal columns}, \\ 
\end{aligned}
\end{equation}

where $\text{diag}(\boldsymbol{b})$ is a diagonal matrix whose diagonal is composed of the vector $\boldsymbol{b}$ of coefficients. The CEN solution for $C_1,...,C_K$ can be found by setting $\widehat{C}_k := \lbrace j \ | \ \widehat{h}_{jk} = 1 \rbrace$, $\forall k$.

\begin{proof}
The objective function for K-means clustering applied to the variables in the matrix $\vec{X}$ is

\begin{equation}
\label{eq:S1-kmeans_var}
    J(\mathcal{C}) = \sum_{k = 1}^K \sum_{j \in C_k} ||\vec{x}_j -  \frac{1}{p_k}\sum_{j \in C_k} \vec{x}_j||_2^2.
\end{equation}

The minimization problem over $\mathcal{C}$ can be rewritten as a minimization over a binary matrix $\vec{H}$ with orthogonal columns, where the solution is

\begin{equation}
\label{eq:S1-kmeans_MF}
\begin{aligned}
   \widehat{\vec{H}} = & \argmin_{\vec{H}} \ ||\vec{X} - \vec{X}\vec{H}(\vec{H}^\top\vec{H})^{-1}\vec{H}^\top ||_2^2 \\
    & \phantom{\text{s.t. }} \vec{H} \in \lbrace 0, 1 \rbrace^{p \times K} \\ 
    & \phantom{\text{s.t. }} \vec{H} \text{ has orthogonal columns}, \\ 
\end{aligned}
\end{equation}
and the estimated clusters $\widehat{C}_k$ are inferred from $\widehat{\vec{H}}$: $\widehat{C}_k := \lbrace j \ | \ \widehat{h}_{jk} = 1 \rbrace$.

In the CEN grouping penalty, K-means clustering is performed on the variables $\vec{x}_j$ weighted by their respective coefficients $b_j$. This means that the input to the K-means problem is not $\vec{X}$ but rather $\vec{X}\text{diag}(\boldsymbol{b})$. Replacing $\vec{X}$ in Eq.~\eqref{eq:S1-kmeans_MF} with $\vec{X}\text{diag}(\boldsymbol{b})$ results in the solution represented in Eq.~\eqref{eq:S1-CEN_MF}.

\end{proof}

\newpage 

\section{Additional Simulation Results}\label{sec:add_sim_results}

\begin{description}

\item[Results for smallest $n$ and highest correlation:] Average out-of-sample performance for $n = 25$ and $\rho = 0.6$, plotted with respect to the average model size $s$. Each point in the plotted lines represents a value of the hyperparameter (HP) $\delta$ or $\lambda$; all other HPs were tuned using $k$-fold cross-validation. The vertical line indicates the true number of active variables, $20$.
\end{description}

\begin{figure}[H]
    \centering
    \includegraphics[scale = 0.4, trim = {0cm 4cm 0cm 0.25cm}, clip]{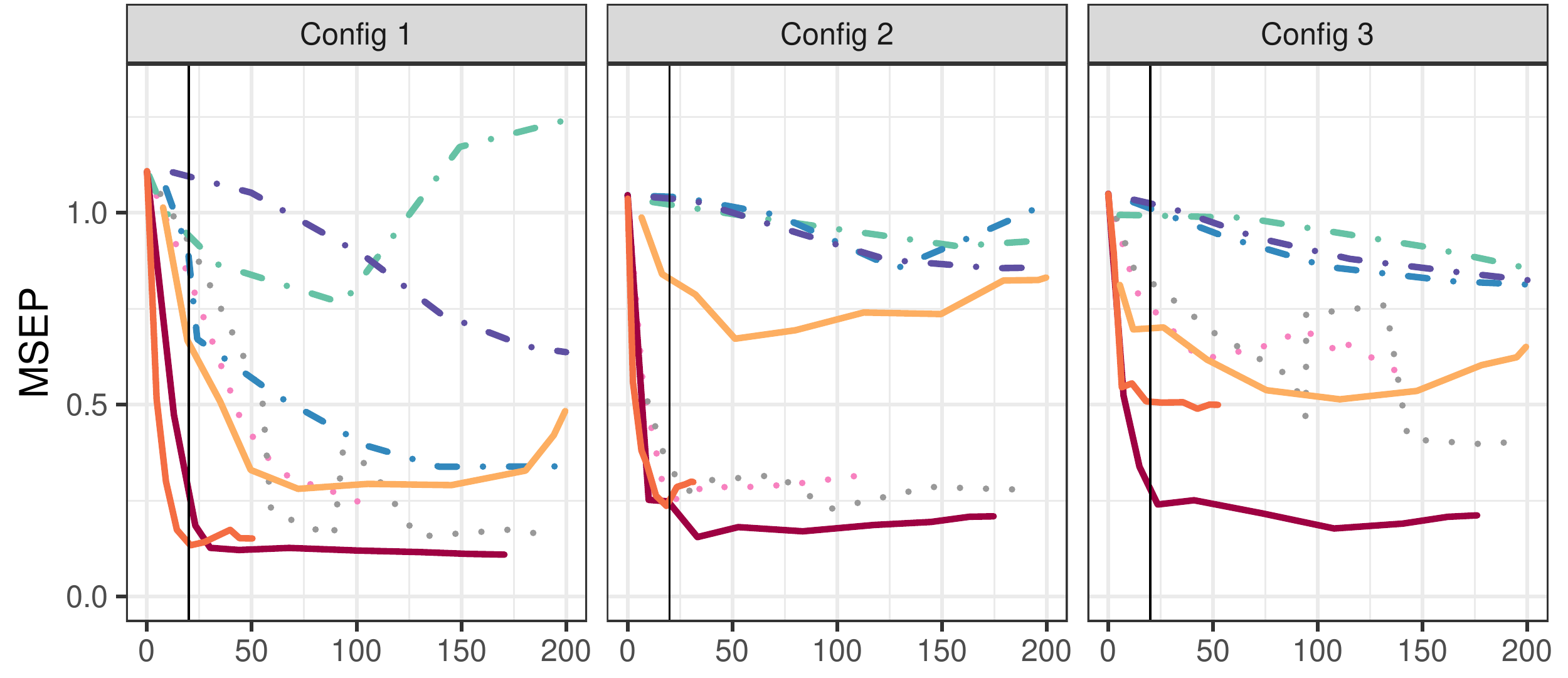}
    
    \includegraphics[scale = 0.4,trim = {0cm 4cm 0cm 0cm}, clip]{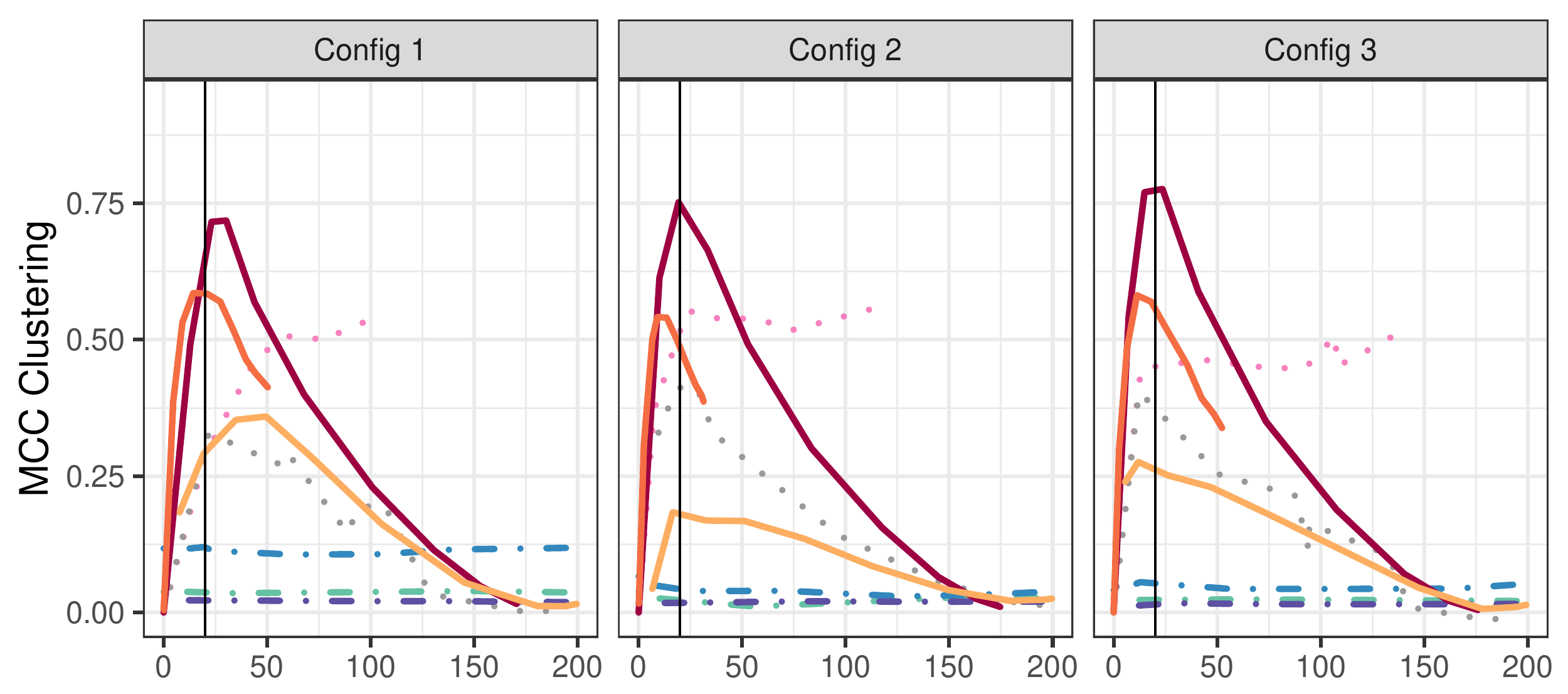}
    
    \includegraphics[scale = 0.4,trim = {0cm 0.5cm 0cm 0cm}, clip]{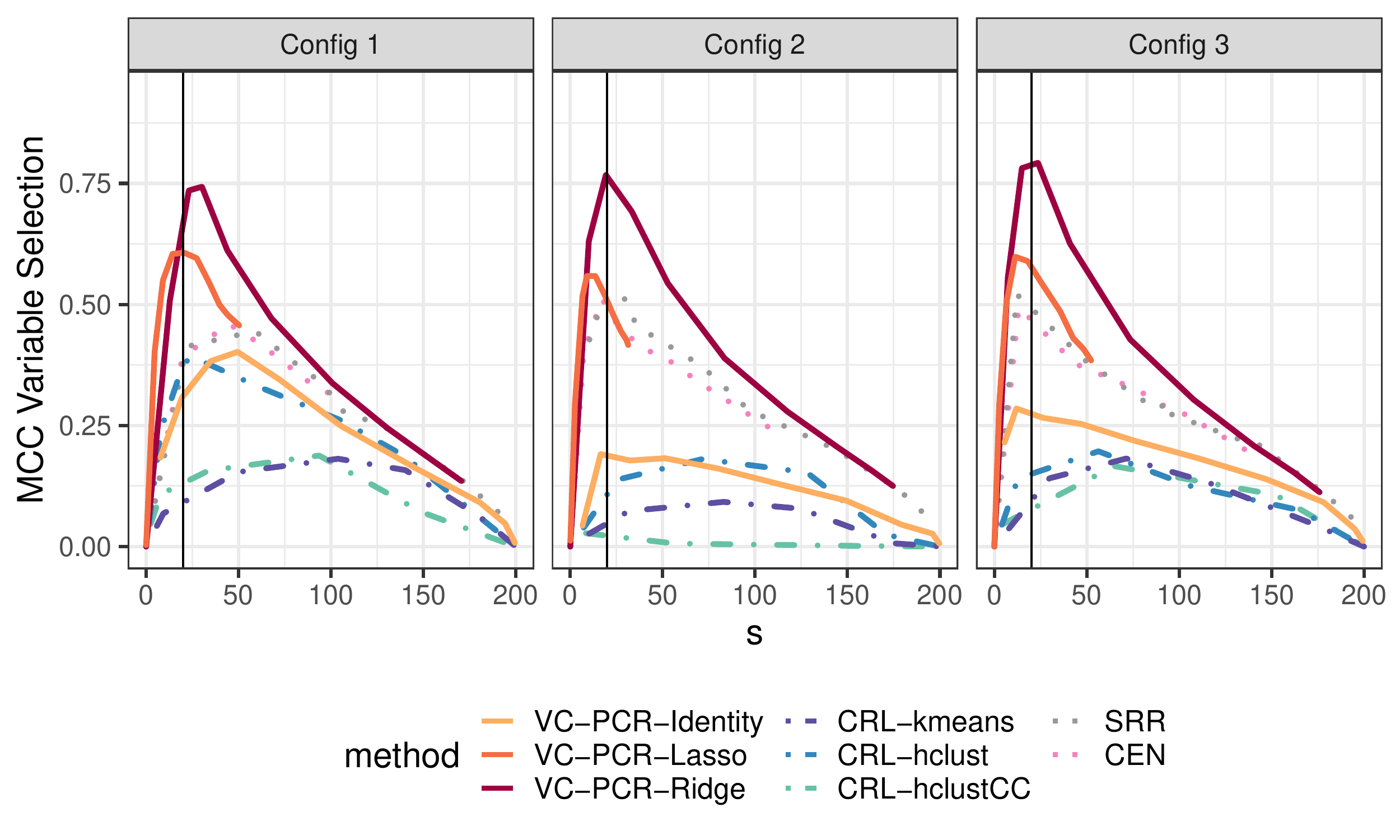}
\caption{}
\end{figure}

\newpage

\begin{description}

\item[Results for largest $n$ and lowest correlation:] 
Average out-of-sample performance for $n = 50$ and $\rho = 0.3$, plotted with respect to the average model size $s$. Each point in the plotted lines represents a value of the hyperparameter (HP) $\delta$ or $\lambda$; all other HPs were tuned using $k$-fold cross-validation. The vertical line indicates the true number of active variables, $20$. 

\end{description}

\begin{figure}[H]
    \centering
    \includegraphics[scale = 0.4, trim = {0cm 4cm 0cm 0.25cm}, clip]{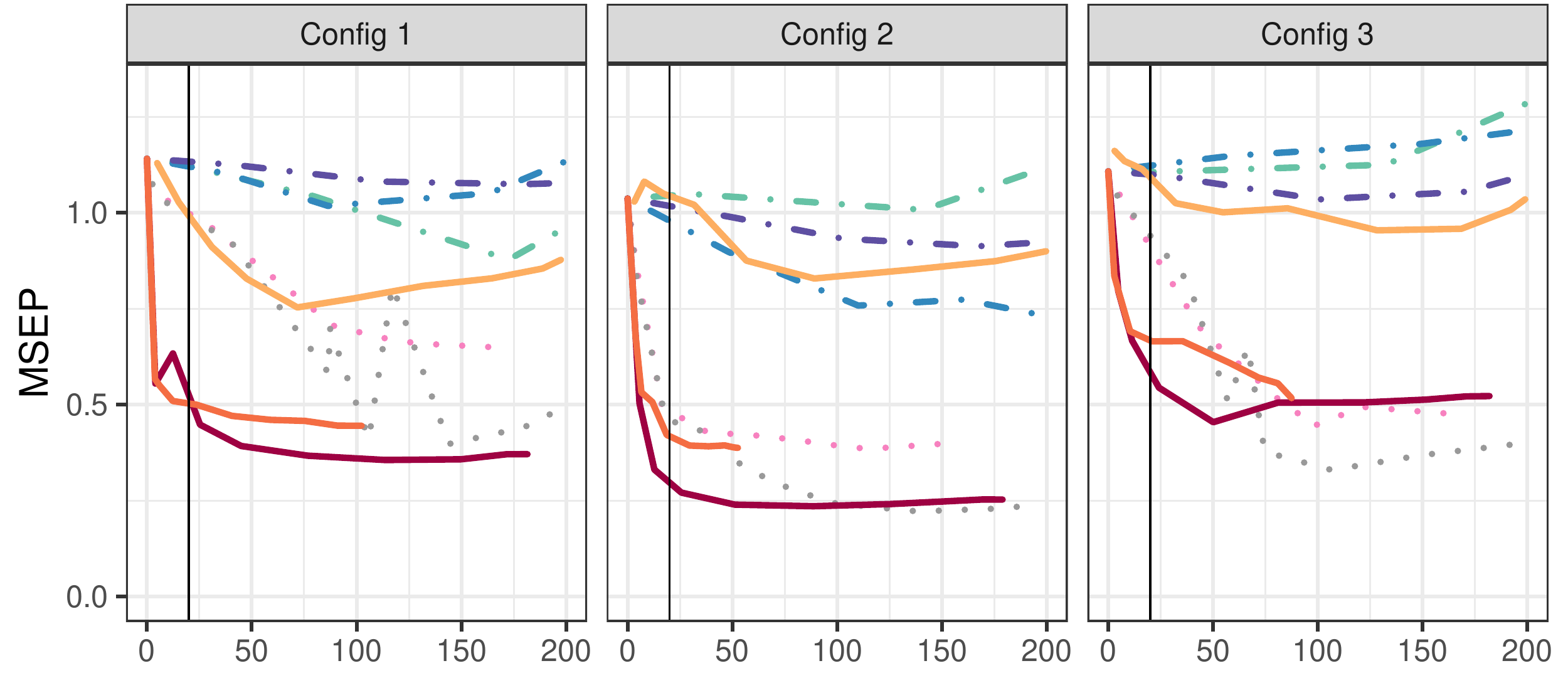}
    
    \includegraphics[scale = 0.4,trim = {0cm 4cm 0cm 0cm}, clip]{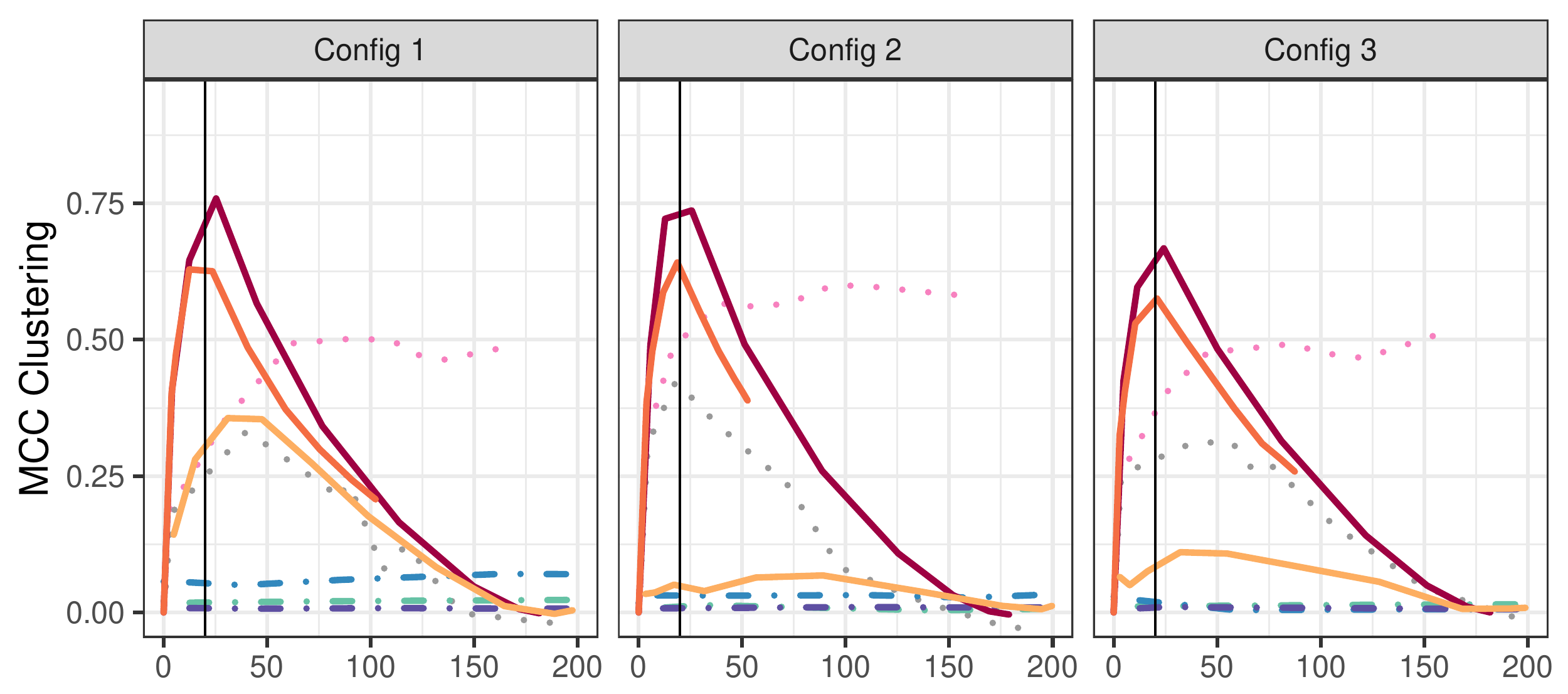}
    
    \includegraphics[scale = 0.4,trim = {0cm 0.5cm 0cm 0cm}, clip]{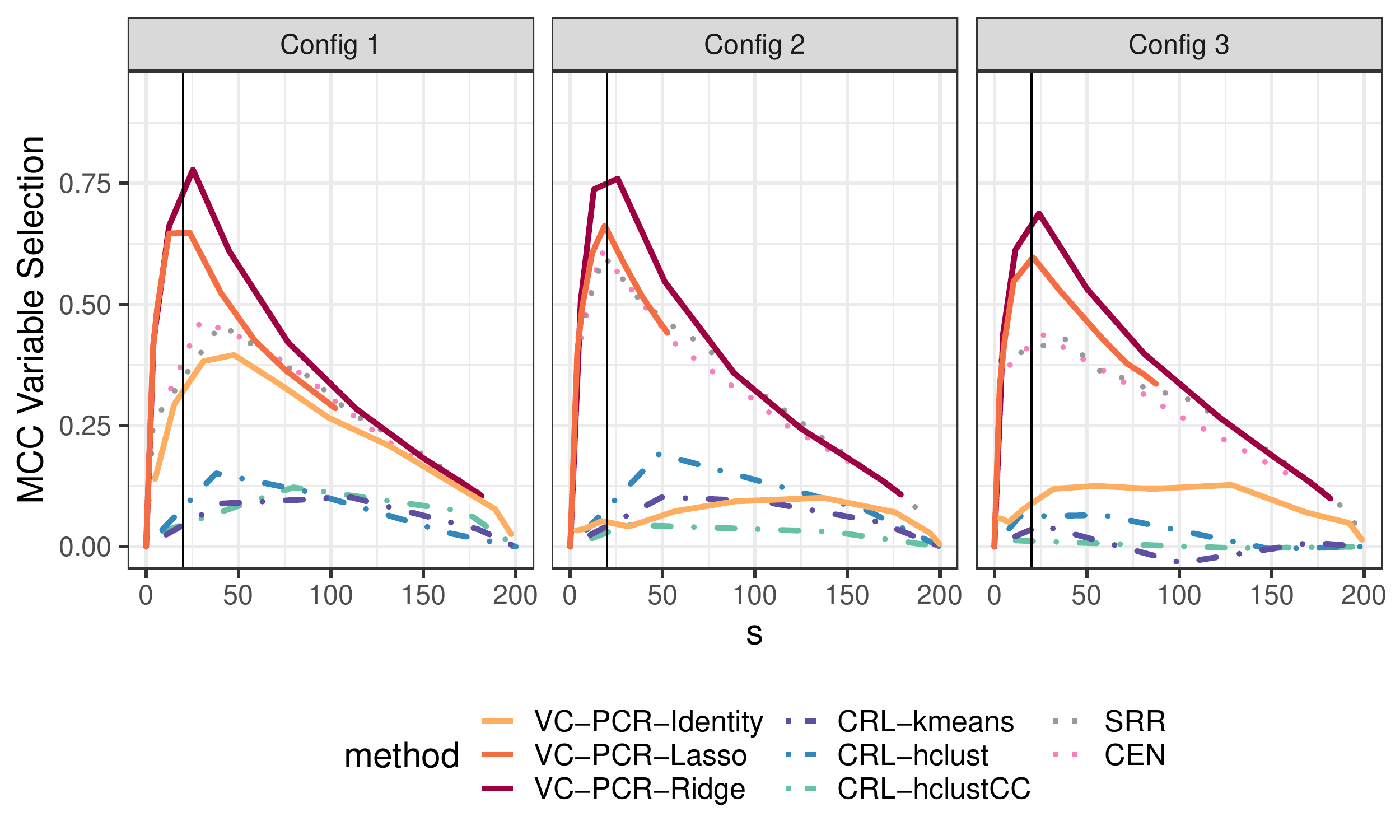}
\caption{}
\end{figure}

\newpage

\begin{description}

\item[Results for smallest $n$ and lowest correlation:] 
Average out-of-sample performance for $n = 25$ and $\rho = 0.3$, plotted with respect to the average model size $s$. Each point in the plotted lines represents a value of the hyperparameter (HP) $\delta$ or $\lambda$; all other HPs were tuned using $k$-fold cross-validation. The vertical line indicates the true number of active variables, $20$. 

\end{description}

\begin{figure}[H]
    \centering
    \includegraphics[scale = 0.4, trim = {0cm 4cm 0cm 0.25cm}, clip]{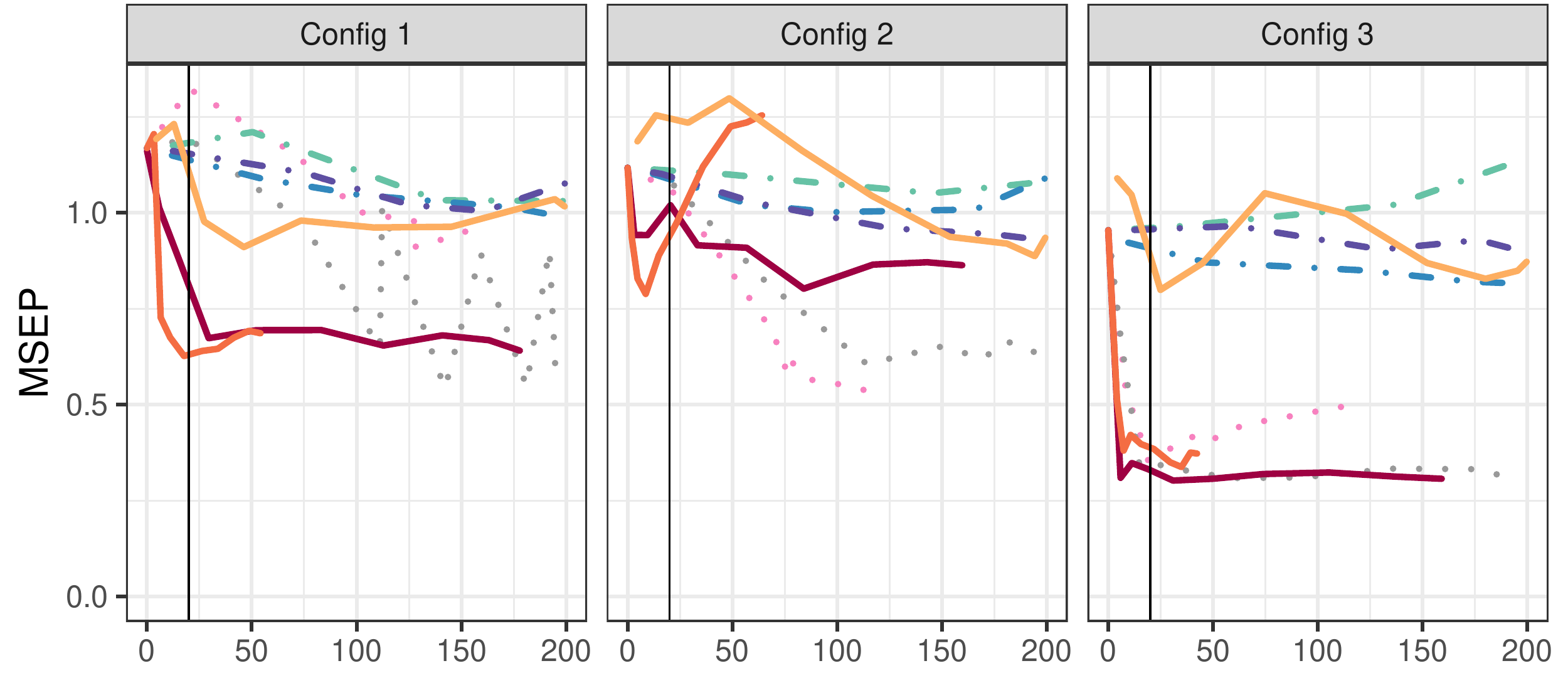}
    
    \includegraphics[scale = 0.4,trim = {0cm 4cm 0cm 0cm}, clip]{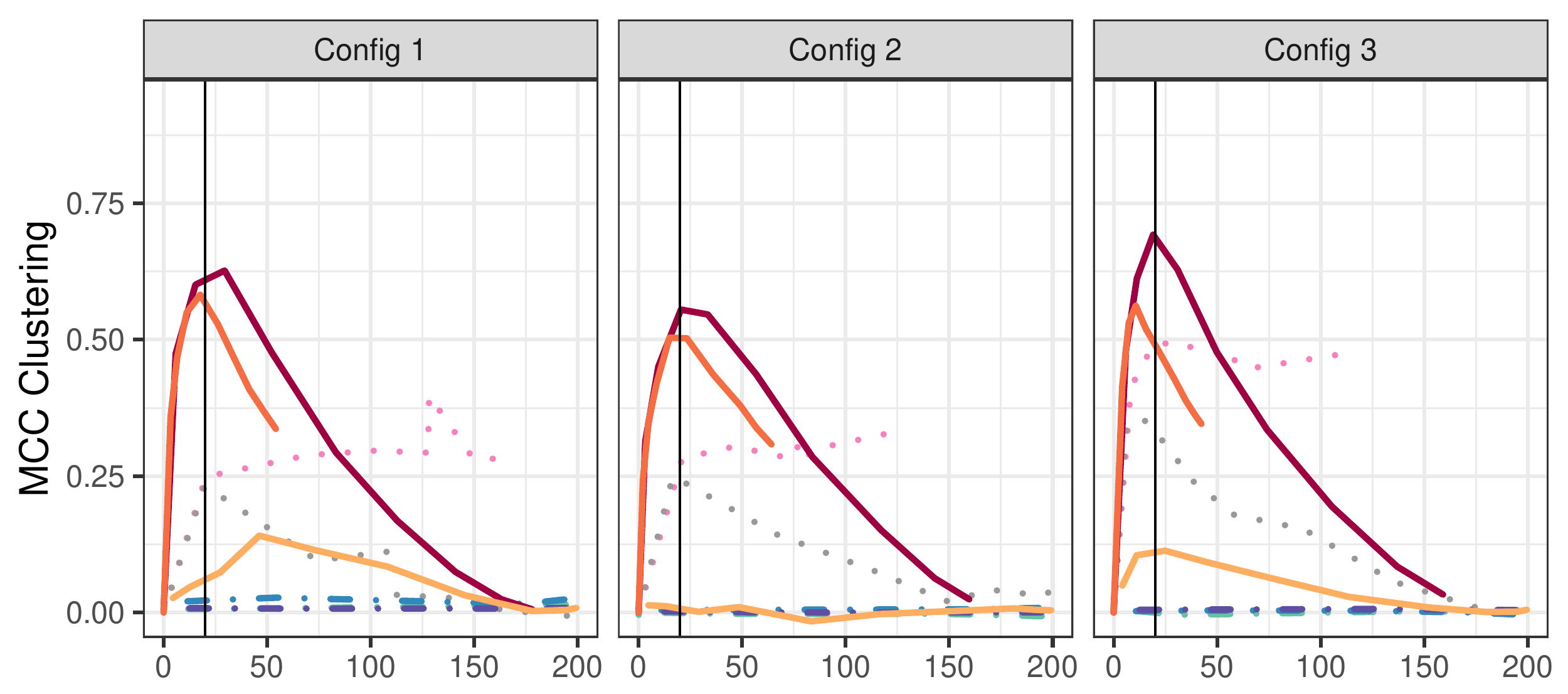}
    
    \includegraphics[scale = 0.4,trim = {0cm 0.5cm 0cm 0cm}, clip]{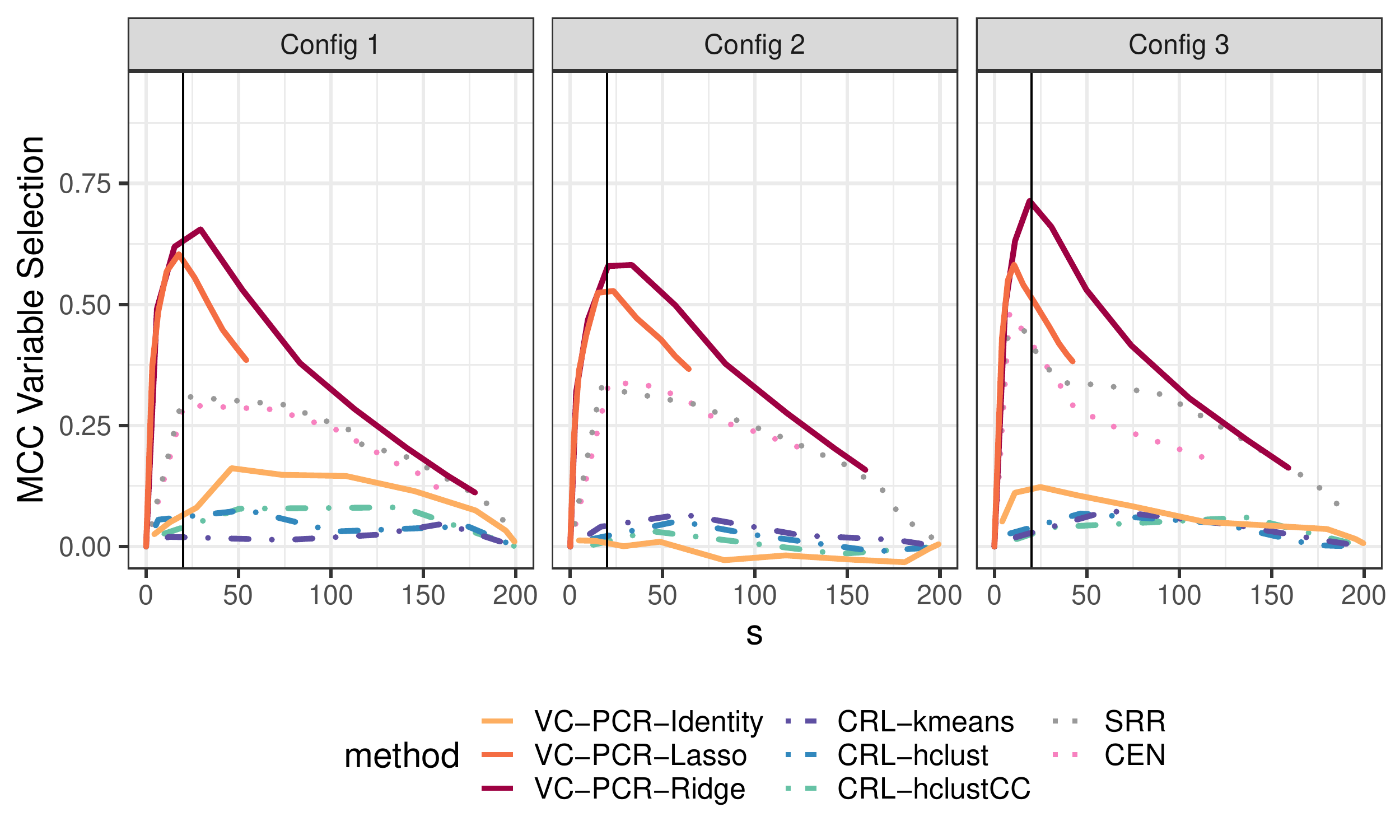}
\caption{}
\end{figure}



\printbibliography



\end{document}